\documentclass{article} 
\usepackage{iclr2027_conference,times}

\iclrfinalcopy

\usepackage{amsmath,amsfonts,bm}

\def\eqref#1{equation~\ref{#1}}

\def\1{\bm{1}}

\DeclareMathAlphabet{\mathsfit}{\encodingdefault}{\sfdefault}{m}{sl}
\SetMathAlphabet{\mathsfit}{bold}{\encodingdefault}{\sfdefault}{bx}{n}

\usepackage{amsmath,amssymb,amsfonts}
\usepackage{graphicx}
\usepackage{textcomp}
\usepackage[table]{xcolor}   
\usepackage{booktabs}
\usepackage{multirow}
\usepackage{enumitem}
\usepackage{array}
\usepackage{tabularx}
\usepackage{longtable}
\usepackage{makecell}
\usepackage{caption}
\usepackage{subcaption}
\usepackage{wrapfig}
\usepackage{placeins}
\usepackage{needspace}
\usepackage{float}
\usepackage{algorithm}
\usepackage{algorithmic}   
\usepackage{xspace}
\usepackage{pifont}
\usepackage{url}
\usepackage{hyperref}

\newcommand{\sysname}{\textsc{Gram}\xspace}

\graphicspath{{figures/}}

\title{\raggedright GRAM: Correcting Frozen Time-Series Foundation Models via Graph-Retrieved Amplitude Memory}

\author{
  \textbf{Xiaoyun Yu}$^{1}$, \textbf{Xiangfei Qiu}$^{1}$, \textbf{Yonggui Huang}$^{2}$, \textbf{Shixiang Tang}$^{3}$, \textbf{Nanqing Dong}$^{4,5}$ \\
  \textbf{Wanli Ouyang}$^{5}$, \textbf{Geguang Pu}$^{1}$, \textbf{Honggang Qi}$^{6}$, \textbf{Jilin Hu}$^{1}$, \textbf{Xi Chen}$^{1}$ \\
  \vspace{-0.5ex} \\
  \begin{tabular}{l}
    $^1$ East China Normal University \\
    $^2$ Peking University \\
    $^3$ The Chinese University of Hong Kong \\
    $^4$ Shanghai Innovation Institute \\
    $^5$ Shanghai Artificial Intelligence Laboratory \\
    $^6$ University of the Chinese Academy of Sciences
  \end{tabular}
}

\begin{document}

\maketitle
\lhead{Preprint} 

\begin{abstract}
Time-series foundation models (TSFMs) enable zero-shot forecasting through large-scale cross-domain pretraining, while retrieval augmentation further improves their performance by leveraging historical information. However, existing methods typically correct TSFM forecasts using the ground-truth futures of similar historical windows, which contain both predictive components already captured by the foundation model and sample-specific random fluctuation that is difficult to transfer. In contrast, recurring systematic model bias within prediction errors more directly characterizes the failure modes of a frozen TSFM and therefore provides more valuable correction signals. Effectively exploiting such model bias, however, poses two challenges: prediction errors at different numerical levels are difficult to compare due to scale differences, and the recurring bias must be extracted from prediction errors contaminated by random fluctuation. To address these challenges, we propose GRAM, a general retrieval-augmented framework for frozen TSFMs. GRAM first introduces an Amplitude Memory Module (AMM) that scales prediction errors by amplitude and aggregates them into retrievable prototypes. It then employs a Prototype Graph Module (PGM) to model relations among prototypes to aggregate consistent bias information while suppressing random fluctuation. During online forecasting, GRAM retrieves and expands prototypes relevant to the current query and generates per-horizon corrections to refine the original TSFM forecast. Experiments across multiple datasets and foundation models demonstrate consistent forecasting improvements.

\end{abstract}

\section{Introduction}
\label{sec:introduction}

Time-series foundation models (TSFMs) are pretrained on large-scale cross-domain corpora~\cite{shi2025timemoe,liu2024timer,goswami2024moment,ekambaram2024ttm,rasul2023lagllama,garza2023timegpt,gruver2023llmtime,jin2024timellm,zhou2023gpt4ts,liu2024moiraimoe,liang2024fmsurvey,wen2022survey}, enabling a single model to transfer to diverse target scenarios for zero-shot forecasting without training a separate predictor for each dataset. This capability is particularly valuable when target-domain data are limited, model updates are costly, or the deployed predictor must remain frozen. However, despite their strong generalization ability, a pretrained TSFM may still exhibit systematic forecasting errors when transferred to a specific target domain~\cite{guo2017calibration,kim2022revin,yao2022wildtime,glahn1972mos}. Effectively correcting TSFM prediction errors and improving forecasting accuracy while keeping the model parameters frozen is therefore of significant research and practical importance.

Recent retrieval-augmented forecasting methods alleviate this problem by exploiting historical data from the target domain~\cite{yang2024timerag,zhou2026stationarity}. These methods retrieve historical windows with similar temporal patterns~\cite{agrawal1993pattern,keogh2002motif} and use the ground-truth future trajectories that follow them as auxiliary evidence. Their implicit assumption is that similar historical patterns correspond to transferable future trajectories. However, this strategy overlooks the forecasting capability that the frozen TSFM already possesses. For a historical window $i$, let $\hat{\mathbf{y}}_i^{\mathrm{tsfm}}$ denote the prediction of the frozen TSFM and $\mathbf{y}_i$ the ground-truth future. Conceptually, the ground-truth future can be expressed as
\[
\mathbf{y}_i
=
\underbrace{\hat{\mathbf{y}}_i^{\mathrm{tsfm}}}_{\text{model prediction}}
+
\underbrace{\mathbf{b}_i}_{\text{model bias}}
+
\underbrace{\boldsymbol{\epsilon}_i}_{\text{sample-specific variation}},
\]
where $\mathbf{b}_i$ denotes a recurring systematic bias under similar historical patterns, and $\boldsymbol{\epsilon}_i$ denotes sample-specific variation that is not reliably transferable. For the current query $q$, the frozen TSFM has already produced its prediction $\hat{\mathbf{y}}_q^{\mathrm{tsfm}}$. Directly retrieving $\mathbf{y}_i$ therefore introduces two components irrelevant to correction. The term $\hat{\mathbf{y}}_i^{\mathrm{tsfm}}$ is already covered by the query's own prediction $\hat{\mathbf{y}}_q^{\mathrm{tsfm}}$, while $\boldsymbol{\epsilon}_i$, as non-transferable random fluctuation, would interfere with the forecast. What is truly valuable for correction is only the recurring model bias $\mathbf{b}_i$.

For non-stationary time series~\cite{liu2022nonstationary}, this problem is particularly pronounced: since the underlying generating mechanism of the series evolves over time, even two historical windows with similar patterns may lead to substantially different futures in level, amplitude, or trend. This motivates us to shift the retrieval target from complete future trajectories to prediction residuals. The residual of a historical window $i$ is defined as
\[
\mathbf{r}_i
=
\mathbf{y}_i-\hat{\mathbf{y}}_i^{\mathrm{tsfm}}
=
\mathbf{b}_i+\boldsymbol{\epsilon}_i.
\]
After removing the model's prediction, the residual retains only the model's forecasting error, directly corresponding to the correction objective. However, a single residual is still not an ideal correction signal: the presence of $\boldsymbol{\epsilon}_i$ means that directly reusing the residual of one historical window may transfer random fluctuation along with the bias. Therefore, it is necessary to aggregate residual evidence from multiple related historical windows to extract the recurring model bias $\mathbf{b}_i$.

However, this process faces two challenges. First, absolute residuals are strongly affected by the local numerical level of the series. Two historical windows may exhibit the same relative prediction bias, yet produce absolute residuals of vastly different magnitudes due to different numerical levels. Standard normalization can only align the global scale of a series and cannot eliminate local numerical differences between windows. Therefore, absolute residuals cannot be directly compared across windows with different numerical levels, and a scale-invariant bias representation is required. Second, the residual of a single historical window contains random fluctuation, making it difficult to reliably identify recurring model bias. Moreover, model bias and historical patterns are not in one-to-one correspondence: different historical patterns may produce similar model bias, and similar historical patterns may also produce different model bias. Useful bias information may therefore be dispersed across multiple distinct but related historical windows, requiring explicit modeling of the relationships among windows to extract recurring model bias.

To address the above problems, we propose \sysname, a retrieval-augmented forecasting framework for frozen TSFMs. \sysname\ constructs a structured bias memory and uses model bias to correct the current forecast, comprising two phases: Offline Memory Construction and Online Forecasting. In the Offline Memory Construction phase, the Amplitude Memory Module (AMM) represents model bias using amplitude scaling error rather than absolute residuals, making bias comparable across historical windows with different numerical levels, and aggregates historical windows into retrievable prototypes. The Prototype Graph Module (PGM) establishes relationships among prototypes, organizing bias information dispersed across different historical windows into a unified structure. In the Online Forecasting phase, \sysname\ retrieves prototypes relevant to the current query and expands along the prototype graph to obtain broader bias information. The Graph Fusion Module (GFM) aggregates the retrieved bias information into per-horizon corrections, which are added to the original forecast of the frozen TSFM. Throughout the entire process, the TSFM and the retrieval encoder remain frozen.

Our contributions are summarized as follows:
\begin{itemize}[leftmargin=*,topsep=2pt,itemsep=2pt]

\item We propose retrieving the prediction residuals of the TSFM on historical windows, so that retrieval retains only the information directly relevant to model correction.

\item We design the Amplitude Memory Module (AMM), which uses amplitude scaling error to characterize the prediction bias of the frozen TSFM under different historical patterns, and organizes related historical windows into retrievable prototypes.

\item We design the Prototype Graph Module (PGM), which establishes associations among prototypes to aggregate bias information dispersed across multiple historical windows.

\item We conduct extensive experiments on seven public benchmark datasets and four time-series foundation models. The results show that \sysname\ achieves the best forecasting performance across different datasets and foundation models. Additionally, the code is available at \url{https://github.com/ecnuabmlab/gram}.

\end{itemize}
\section{Related Work}
\label{sec:related}

\subsection{Time-Series Forecasting with Foundation Models}
\label{sec:related-tsfm}

Time-series forecasting has progressed from statistical
models~\cite{box2015arima} to deep architectures that capture
long-range, cross-variable, frequency or multi-scale
patterns~\cite{salinas2020deepar,lim2021tft,oreshkin2020nbeats,challu2023nhits,zhou2022fedformer,liu2022pyraformer,zhang2023crossformer,chen2023tsmixer}; all of these models are trained per dataset and
depend on target labels. TSFMs instead pretrain once on large
cross-domain corpora and forecast zero-shot on unseen
domains,
removing the dependence on target labels. Parameter-efficient
adaptation further reduces the number of updated parameters but still
requires target-specific optimization. In contrast,
\sysname\ keeps the TSFM frozen and corrects its systematic scaling
errors through an external memory.

\subsection{Retrieval-Augmented Forecasting}
\label{sec:related-rag}

Retrieval-augmented methods complement parametric models with
non-parametric information selected from an external
corpus~\cite{johnson2019faiss}. The
principle originates in language
modeling~\cite{lewis2020rag,guu2020realm,izacard2022atlas,karpukhin2020dpr,khandelwal2020knnlm,borgeaud2022retro,gao2023ragsurvey} and has
recently been extended to time-series forecasting. Existing methods
differ mainly in how the retrieved evidence is incorporated.
TimeRAG~\cite{yang2024timerag} builds a knowledge base of time-series
segments and couples retrieval with a large language model;
TimeRAF~\cite{zhang2025timeraf} trains a bi-encoder that retrieves
historical windows and uses their futures as auxiliary signals for the
forecast; TS-RAG~\cite{ning2025tsrag} augments a frozen TSFM with a
retrieval knowledge base and fuses the retrieved patterns with the
TSFM representation through an attention-based mixer; related
designs adopt similar retrieve-then-fuse pipelines~\cite{han2025raft}.
These methods demonstrate that target-domain history can improve a
frozen forecaster, but they generally treat retrieved windows as
independent items and use similarity as the primary relevance signal.
\sysname\ instead retrieves the scaling error that the frozen TSFM
leaves on a pattern and reads it through a prototype graph: prototypes
are connected by bias and shape relations, and the retrieved subgraph
is processed by a trainable relational message-passing head whose
output is gated by the evidence behind each retrieved
prototype.

\section{Method}
\label{sec:method}

We study zero-shot forecasting with a frozen time-series foundation model (TSFM). The input is a context window $\mathbf x_q$ of recent observations, and the goal is to predict the next $H$ steps $\mathbf y_q$ without using any label of the target series. Two components stay frozen throughout: a TSFM $F$ that produces the zero-shot forecast $\widehat{\mathbf y}_q^{\mathrm{tsfm}}=F(\mathbf x_q)$, and a retrieval encoder $E$ that maps any window to a unit-norm embedding; we write $e_q$ for the query and $e_1,\ldots,e_n$ for the historical windows. Retrieval-augmented forecasting keeps this pipeline but reuses the ground-truth futures that follow the retrieved windows. Such a future, however, contains the model's own prediction and non-transferable sample-specific fluctuation on top of the recurring model bias, so reusing it transfers both a component already covered by the query's own forecast and fluctuation that interferes with it; only the recurring bias is useful for correction. \sysname\ keeps the same retrieval pipeline but changes the target of retrieval accordingly: each prototype stores, for a recurring pattern, the amplitude of the systematic error of the TSFM, and a trainable read-out turns the retrieved subgraph into an error component added to the frozen forecast.

Let $t_q$ denote the last observed step of the query window and $S$ the number of steps that define the recent level. The recent level, the predicted excursion and the realized excursion are
\begin{equation}
m_q \;=\; \frac{1}{S}\sum_{t=t_q-S+1}^{t_q} x_{q,t},
\qquad
\mathrm{exc}_q(h) \;=\; \widehat y_q^{\mathrm{tsfm}}(h) - m_q,
\qquad
\mathrm{res}_q(h) \;=\; y_q(h) - m_q,
\label{eq:exc-res}
\end{equation}
for $h\in\{1,\ldots,H\}$. Their ratio $\gamma_q(h)=\mathrm{res}_q(h)/\mathrm{exc}_q(h)$ yields the factorization $y_q(h)-\widehat y_q^{\mathrm{tsfm}}(h)=\mathrm{res}_q(h)-\mathrm{exc}_q(h)=\bigl(\gamma_q(h)-1\bigr)\,\mathrm{exc}_q(h)$: the first factor is dimensionless and states how the TSFM mis-scales the departure, while the second carries the local numerical level. Storing the first factor and re-applying it to the query's own excursion is what makes the memory scale-invariant.

\subsection{Overview}
\label{sec:method-overview}
\begin{figure*}[t]
 \centering
 \includegraphics[width=0.99\textwidth]{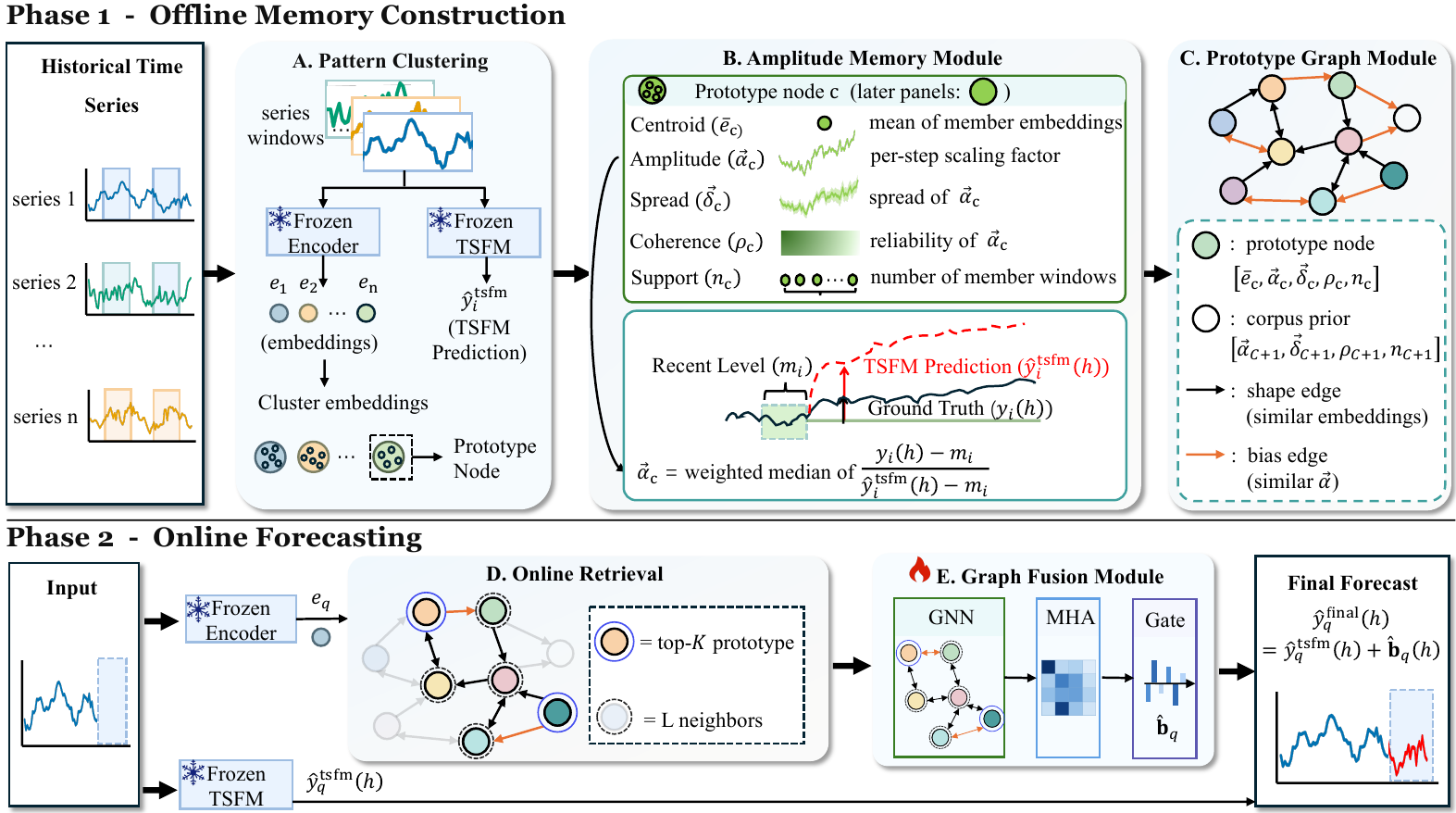}
 \caption{Overview of \sysname. Phase~1 (Offline Memory Construction): clustering (A) groups window embeddings into prototypes, the amplitude memory (B) stores each prototype's centroid, amplitude, spread, coherence and support, and the prototype graph (C) links them by bias and shape edges. Phase~2 (Online Forecasting): retrieval (D) grows the top-$K$ prototypes into a subgraph and attaches the anchor node, and graph fusion (E), the only trainable module, outputs the error component added to the frozen forecast.}
 \label{fig:architecture}
\end{figure*}

Figure~\ref{fig:architecture} shows the pipeline. Offline, pattern clustering (A) groups the window embeddings into $C$ prototypes, the amplitude memory (B) stores a prototype record for each of them, and the prototype graph (C) links them by the bias and shape relations (Section~\ref{sec:method-offline}). Online, retrieval (D) grows the top-$K$ prototypes into a subgraph and attaches an anchor node, and the graph fusion module (E), the only trainable component, turns the subgraph into an error component $\hat{\mathbf{b}}_q(h)$ that the final forecast adds to the frozen prediction (Section~\ref{sec:method-online}). Setting $\hat{\mathbf{b}}_q(h)\equiv0$ recovers the frozen TSFM exactly.

\subsection{Offline Memory Construction}
\label{sec:method-offline}

The offline phase distills the target history into two artifacts: the amplitude memory, which records how the TSFM errs on each pattern, and the prototype graph, which records which patterns err alike. Amplitudes rather than absolute residuals keep the memory dimensionless and reusable across corpora, and links between patterns that err alike let evidence recorded on one pattern be reused on another. The phase runs once per dataset, TSFM and horizon, and only on the memory-construction segment, so it never sees a test label.

\subsubsection{Pattern Clustering}
\label{sec:method-offline-clustering}

Module~A runs the two frozen components over every window of the target history, producing the embeddings $e_1,\ldots,e_n$ and the forecasts $\hat y_1^{\mathrm{tsfm}},\ldots,\hat y_n^{\mathrm{tsfm}}$, and groups the embeddings into $C$ prototype nodes~\cite{sculley2010kmeans,snell2017prototypical}. The centroid $\bar e_c$ of prototype $c$ is the unit-normalized mean of its member embeddings; it is the identity of the node and the key used for retrieval. A window is assigned to the closest centroid, $z_i=\arg\min_c\|e_i-\bar e_c\|_2$, and each cluster records its support $n_c$, the number of member windows behind it. The segment split and the values of $C$ and $K$ are reported in Appendix~\ref{app:reproducibility}.

\subsubsection{Amplitude Memory}
\label{sec:method-offline-prototype}

The amplitude memory records, for every cluster, how the TSFM mis-scales the excursions of its member windows. For a member window $i$ of cluster $c$, the per-step ratio $\gamma_i(h)=\mathrm{res}_i(h)/\mathrm{exc}_i(h)$ is the factor between the realized and predicted excursions at step $h$. The cluster aggregates these ratios into an amplitude vector $\vec\alpha_c\in\mathbb R^{H}$ and a spread vector $\vec\delta_c\in\mathbb R^{H}$,
\begin{align}
\alpha_c(h) &\;=\; \operatorname{med}_{i:z_i=c}\bigl[\gamma_i(h);\;\omega_i(h)=|\mathrm{exc}_i(h)|\bigr],\label{eq:amp}\\
\delta_c(h) &\;=\; \frac{\sum_{i:z_i=c} \omega_i(h)\,\bigl|\gamma_i(h)-\alpha_c(h)\bigr|}{\sum_{i:z_i=c} \omega_i(h)},\label{eq:spread}
\end{align}
where $\operatorname{med}$ denotes the $\omega_i(h)$-weighted median; the weight down-weights windows whose excursion is close to zero, because they have no departure to rescale. We call $\alpha_c(h)-1$ the \emph{amplitude scaling error} of pattern $c$ at lead $h$: in the decomposition of the introduction, $\bigl(\alpha_c(h)-1\bigr)\,\mathrm{exc}_i(h)$ is the recurring bias $\mathbf b_i$ of a member window expressed relative to its own excursion, while $\vec\delta_c$ summarizes the sample-specific component $\boldsymbol\epsilon_i$. The cluster also records a coherence score
\begin{equation}
\rho_c \;=\; \frac{1}{H}\sum_{h=1}^{H}
\frac{\alpha_c(h)^{2}}{\alpha_c(h)^{2}+\delta_c(h)^{2}+\varepsilon_0},
\label{eq:coherence}
\end{equation}
which is close to one when the member windows agree on how the TSFM erred on the pattern. Each prototype therefore carries the five-tuple $\bigl[\bar e_c,\vec\alpha_c,\vec\delta_c,\rho_c,n_c\bigr]$, of which the centroid serves as the identity of the node for retrieval and the remaining attributes are consumed by the read-out.

\subsubsection{Prototype Graph}
\label{sec:method-offline-graph}

The prototype graph connects the $C$ prototypes with two relations, each prototype keeping its strongest neighbors under each relation; the out-degree is a hyper-parameter reported in the experiments. For the \emph{bias relation}, we subtract one from each amplitude vector, forming the scaling-error curve $\vec\alpha_c-\mathbf 1$, and $\ell_2$-normalize it; the bias similarity is the cosine of these curves, and a bias edge requires the neighbor's support to exceed a floor $n_{\min}$, without which low-support prototypes could be linked by noise. For the \emph{shape relation}, the similarity is the cosine of the centroids $\bar e_c$; since a prototype with very small support has an amplitude close to one, because the weighted median over few members shrinks toward one, its shape neighbors provide the usable evidence for the read-out. The bias relation states that two prototypes are mis-scaled in the same way, which is the property that allows an amplitude correction to transfer between them. Finally, a corpus prior summarizes the whole library as the support-weighted average,
\begin{equation}
\vec\alpha_{C+1}=\frac{\sum_c n_c\,\vec\alpha_c}{\sum_c n_c},\qquad
\vec\delta_{C+1}=\frac{\sum_c n_c\,\vec\delta_c}{\sum_c n_c},\qquad
\rho_{C+1}=\frac{\sum_c n_c\,\rho_c}{\sum_c n_c},\qquad
n_{C+1}=\sum_c n_c,
\label{eq:corpus-prior}
\end{equation}
attached to the graph as an embedding-free node: it carries no centroid, is excluded from message passing, and only votes in the anchor aggregation of the online phase.

\subsection{Online Forecasting}
\label{sec:method-online}

The online phase serves one query at a time without accessing any target label. The aggregated retrieval answer is attached as an anchor node and enters the correction with a fixed weight, so the read-out always contains the simplest retrieval baseline and every gain over it has to come from the graph; the trained part is gated by the evidence behind the vote, so it shrinks when the retrieved prototypes disagree.

\subsubsection{Online Retrieval}
\label{sec:method-online-retrieval}

Retrieval compares $e_q$ with the centroids, keeps the $K$ best prototypes as seeds $\mathcal S_q=\{c_1,\ldots,c_K\}$, and grows them one hop along the two relations, keeping the $L$ strongest neighbors of each seed. The stored record of every node in this subgraph is loaded from the amplitude memory together with the seed similarities $\mathrm{sim}_{q,c}$, and the aggregated answer over the seeds and the corpus prior is attached as the anchor node:
\begin{equation}
\tilde\alpha_q(h) \;=\; \sum_{j\in\mathcal S_q\cup\{C+1\}}
\tilde w_j\,\alpha_j(h),\qquad
\tilde w_j \;\propto\; \operatorname{softmax}_{j\in\mathcal S_q\cup\{C+1\}}(\mathrm{sim}_{q,j})\,\rho_j,
\label{eq:anchor}
\end{equation}
where the corpus prior, which carries no centroid, is given the neutral similarity $\mathrm{sim}_{q,C+1}=\frac{1}{K}\sum_{j\in\mathcal S_q}\mathrm{sim}_{q,j}$. The anchor carries the aggregated record in the same five-tuple format (Appendix~\ref{app:gfm}); since it joins the subgraph without any edge, attending to it alone reproduces the fixed-fraction baseline that this curve defines.

\subsubsection{Graph Fusion}
\label{sec:method-online-fusion}

The graph fusion module maps the subgraph to the per-horizon error component $\hat{\mathbf{b}}_q$ through three blocks, GNN, MHA and Gate, as shown in module~E of Figure~\ref{fig:architecture}. It is fitted once on a large multi-domain corpus by minimizing the error of the corrected forecast, under the same frozen-TSFM contract: only the head is trained, the TSFM and the retrieval encoder remain frozen, and no label of the evaluation domains is used (Appendix~\ref{app:gfm}). Each node is featurized with its stored curves and scalar statistics, and a relational message-passing network with a separate projection per relation propagates them over the two relations (Appendix~\ref{app:gfm}); the anchor is excluded from message passing because it carries no edges. The query is encoded from scale statistics of its own excursion (Appendix~\ref{app:gfm}); a multi-head attention layer~\cite{vaswani2017attention} takes this vector as the only query and the propagated node representations as keys and values, and returns an attention-pooled vector together with the attention weights $\boldsymbol\pi_q$, one per subgraph node. A two-layer head maps the pooled and query features to a per-lead gain $g(h)\ge0$ that weights the voted scaling, and its last linear layer is initialized so that $g\equiv1$ at the start of training: the head begins by passing the vote through unweighted and learns only the per-lead departures from it.

\paragraph{Gate block.}
The attention weights define a voted curve,
\begin{equation}
v(h) \;=\; \sum_c \pi_{q,c}\,\bigl(\alpha_c(h)-1\bigr),\label{eq:vote}
\end{equation}
a parameter-free, attention-weighted scaling error. The gate caps the scaling with the evidence behind the vote,
\begin{equation}
\phi(h) \;=\; \sigma\!\Bigl( b + \kappa_0\!\!\sum_c \pi_{q,c}\,\mathrm{rel}(n_c)
 + \kappa_1\!\!\sum_c \pi_{q,c}\,\rho_c
 - \kappa_2\!\!\sum_c \pi_{q,c}\,\delta_c(h)
 - \kappa_3\!\!\sum_c \pi_{q,c}\,\bigl|\alpha_c(h)-1-v(h)\bigr|\Bigr),
\label{eq:ceiling}
\end{equation}
where $\mathrm{rel}(n_c)=\log(1+n_c)-\log(1+\mathrm{med}_{c'}n_{c'})$ is the support relative to the median of the $C$ prototypes, the gains $\kappa_i=\mathrm{softplus}(\cdot)$ are non-negative, and the sums run over the nodes of the subgraph; Appendix~\ref{app:gfm} details each term. The signs are fixed: larger support and coherence can raise the ceiling, while larger spread or node disagreement must lower it. Since $\phi(h)$ lies in $(0,1)$ and $v(h)$ is bounded by the stored amplitudes, the gate can shrink the trained part of the correction but cannot change its sign, while the two fixed terms hold it at the level of the retrieved answer and of the corpus.

\subsubsection{Final Forecast}
\label{sec:method-online-final}

The final forecast adds the error component to the frozen prediction,
\begin{align}
\widehat y_q^{\mathrm{final}}(h) &\;=\; \widehat y_q^{\mathrm{tsfm}}(h) + \hat{\mathbf{b}}_q(h),\qquad h=1,\ldots,H,\label{eq:final-forecast}\\
\hat{\mathbf{b}}_q(h) &\;=\; \bigl(a(h)-1\bigr)\,\mathrm{exc}_q(h),\label{eq:correction}\\
a(h) &\;=\; 1 + \mu\bigl(\tilde\alpha_q(h)-1\bigr)
+ \nu\Bigl[\bigl(\alpha_{C+1}(h)-1\bigr) + \phi(h)\,g(h)\,v(h)\Bigr].\label{eq:ratio}
\end{align}
The ratio draws on three sources, of which the first two are fixed before
training and carry the constants $\mu$ and $\nu$ (Appendix~\ref{app:gfm}). The
pooled retrieval answer of Eq.~\eqref{eq:anchor}, scaled by the anchor weight
$\mu$, keeps the simplest retrieval baseline as a standing component of the
correction; the corpus prior of Eq.~\eqref{eq:corpus-prior}, scaled by the
residual weight $\nu$, supplies the library-wide scaling error, so that a quiet
neighbourhood still receives a conservative correction. What training adds is the
last term: the voted scaling of Eq.~\eqref{eq:vote}, capped by the gate
$\phi(h)$ and reweighted per lead by the gain $g(h)$. Every prototype therefore
contributes $\bigl(\alpha_c(h)-1\bigr)\mathrm{exc}_q(h)$, that is, how much it
says the TSFM overstated ($\alpha_c(h)<1$) or understated ($\alpha_c(h)>1$) the
departure at step $h$, and the anchor and the corpus prior enter in the same
units. Read against the decomposition of the introduction, $\hat{\mathbf{b}}_q$
is an estimate of the recurring bias $\mathbf{b}_q$ whose trained part is gated
by the evidence behind it. Setting $\hat{\mathbf{b}}_q(h)\equiv0$ recovers
$\widehat y_q^{\mathrm{tsfm}}$ exactly, so the frozen TSFM is the identity point
of the correction.
\section{Experiments}
\label{sec:experiments}

\paragraph{Datasets.}
We evaluate on seven widely used benchmarks: ETTh1, ETTh2, ETTm1, and
ETTm2~\cite{zhou2021informer}, together with Weather, Electricity, and
Exchange Rate~\cite{wu2023timesnet}, following their canonical
chronological splits. Dataset statistics and the exact split boundaries
are reported in Appendix~\ref{app:datasets}.

\paragraph{TSFMs.}
\sysname\ is attached to four TSFMs with different architectures and
pretraining corpora: Chronos-Bolt~\cite{ansari2024chronos},
Sundial~\cite{liu2025sundial}, TimesFM~\cite{das2024timesfm}, and
Moirai~\cite{woo2024moirai}. All TSFMs remain frozen, and retrieval
embeddings are computed by a frozen Chronos encoder shared across
TSFMs, so a single correction pipeline is applied unchanged to every model.

\paragraph{Setup.}
The context length is fixed to 512, and the forecast horizon is
$H\in\{96,192,336,720\}$. Each channel is forecast independently. For
each dataset--horizon pair and each TSFM, the official training split is
divided chronologically into a memory portion and a calibration portion,
separated from the test split by a horizon-length gap; the Amplitude
Memory Module (AMM) and the Prototype Graph Module (PGM) are built only
from these observed portions, and the Graph Fusion Module (GFM) is
fitted once on the LOTSA corpus~\cite{woo2024moirai} and then frozen for all
target benchmarks and TSFMs. Only the GFM is trained; the TSFM and the
retrieval encoder stay frozen, and no label of the evaluation domains is
used. Since the only fitted part is fixed on an unrelated corpus before the test
portion of a target cell is read, every comparison reported below is zero-shot with
respect to the evaluation domains. The complete protocol, with the
hyper-parameters and the machine used, is reported in Appendix~\ref{app:datasets}.

\paragraph{Baselines and metrics.}
The baseline is the frozen TSFM itself: for each of the four models we compare
its zero-shot forecast with the same forecast corrected by \sysname, on the same
dataset and horizon, so that every comparison below is a like-for-like one between
\emph{Ori} and \emph{+\sysname}. We also report LoRA fine-tuned versions of the
four TSFMs~\cite{hu2021lora}, which show whether the correction survives once the
model has been adapted to the target domain. To ensure consistency with previous
studies, we adopt Mean Squared Error (MSE) and Mean Absolute Error (MAE) as
evaluation metrics, always comparing MSE with MSE and MAE with MAE; lower is better
in both. Their implementation details are reported in
Appendix~\ref{app:baselines}.

\subsection{Main Results}
\label{sec:exp-main}

\begin{table*}[t]
  \centering
  \caption{Zero-shot forecasting comparison between frozen foundation models and the same models corrected by \sysname\ across seven datasets. For each TSFM, \emph{Ori} is the bare frozen model and \emph{+\sysname} the corrected forecast; MSE is compared with MSE and MAE with MAE, and \textbf{\textcolor{red}{red}} marks the better of the two. Results are averaged over horizons $H\in\{96,192,336,720\}$.}
  \resizebox{\textwidth}{!}{
    \begin{tabular}{c|cccc|cccc|cccc|cccc}
    \toprule
    \multirow{2}[4]{*}{Models} & \multicolumn{4}{c|}{\textbf{Sundial}} & \multicolumn{4}{c|}{\textbf{Chronos}} & \multicolumn{4}{c|}{\textbf{TimesFM}} & \multicolumn{4}{c}{\textbf{Moirai}} \\
    \cmidrule{2-17}
     & \multicolumn{2}{c}{Ori} & \multicolumn{2}{c|}{+\sysname\ (Ours)} & \multicolumn{2}{c}{Ori} & \multicolumn{2}{c|}{+\sysname\ (Ours)} & \multicolumn{2}{c}{Ori} & \multicolumn{2}{c|}{+\sysname\ (Ours)} & \multicolumn{2}{c}{Ori} & \multicolumn{2}{c}{+\sysname\ (Ours)} \\
    \midrule
    Metric & MSE & MAE & MSE & MAE & MSE & MAE & MSE & MAE & MSE & MAE & MSE & MAE & MSE & MAE & MSE & MAE \\
    \midrule
    ETTh1    & 0.397 & 0.412 & \textbf{\textcolor{red}{0.389}} & \textbf{\textcolor{red}{0.408}} & 0.448 & 0.418 & \textbf{\textcolor{red}{0.436}} & \textbf{\textcolor{red}{0.414}} & 0.421 & 0.409 & \textbf{\textcolor{red}{0.417}} & \textbf{\textcolor{red}{0.407}} & 0.620 & 0.504 & \textbf{\textcolor{red}{0.530}} & \textbf{\textcolor{red}{0.473}} \\
    \midrule
    ETTh2    & 0.343 & 0.383 & \textbf{\textcolor{red}{0.339}} & \textbf{\textcolor{red}{0.379}} & 0.367 & 0.380 & \textbf{\textcolor{red}{0.362}} & \textbf{\textcolor{red}{0.377}} & 0.372 & 0.388 & \textbf{\textcolor{red}{0.369}} & \textbf{\textcolor{red}{0.385}} & 0.451 & 0.438 & \textbf{\textcolor{red}{0.405}} & \textbf{\textcolor{red}{0.415}} \\
    \midrule
    ETTm1    & 0.366 & 0.370 & \textbf{\textcolor{red}{0.360}} & \textbf{\textcolor{red}{0.367}} & 0.421 & 0.383 & \textbf{\textcolor{red}{0.411}} & \textbf{\textcolor{red}{0.379}} & 0.381 & 0.372 & \textbf{\textcolor{red}{0.378}} & \textbf{\textcolor{red}{0.370}} & 0.740 & 0.523 & \textbf{\textcolor{red}{0.617}} & \textbf{\textcolor{red}{0.485}} \\
    \midrule
    ETTm2    & 0.275 & 0.321 & \textbf{\textcolor{red}{0.264}} & \textbf{\textcolor{red}{0.313}} & 0.291 & 0.317 & \textbf{\textcolor{red}{0.287}} & \textbf{\textcolor{red}{0.315}} & 0.291 & 0.318 & \textbf{\textcolor{red}{0.285}} & \textbf{\textcolor{red}{0.315}} & 0.404 & 0.388 & \textbf{\textcolor{red}{0.344}} & \textbf{\textcolor{red}{0.360}} \\
    \midrule
    ECL      & 0.148 & 0.243 & \textbf{\textcolor{red}{0.147}} & \textbf{\textcolor{red}{0.242}} & 0.159 & 0.244 & \textbf{\textcolor{red}{0.159}} & \textbf{\textcolor{red}{0.244}} & 0.164 & 0.249 & \textbf{\textcolor{red}{0.164}} & \textbf{\textcolor{red}{0.249}} & 0.520 & 0.446 & \textbf{\textcolor{red}{0.391}} & \textbf{\textcolor{red}{0.407}} \\
    \midrule
    Weather  & 0.244 & 0.275 & \textbf{\textcolor{red}{0.236}} & \textbf{\textcolor{red}{0.269}} & 0.263 & 0.276 & \textbf{\textcolor{red}{0.257}} & \textbf{\textcolor{red}{0.273}} & 0.263 & 0.268 & \textbf{\textcolor{red}{0.253}} & \textbf{\textcolor{red}{0.264}} & 0.374 & 0.333 & \textbf{\textcolor{red}{0.315}} & \textbf{\textcolor{red}{0.310}} \\
    \midrule
    Exchange & 0.558 & 0.499 & \textbf{\textcolor{red}{0.439}} & \textbf{\textcolor{red}{0.440}} & 0.371 & 0.412 & \textbf{\textcolor{red}{0.358}} & \textbf{\textcolor{red}{0.402}} & 0.404 & 0.420 & \textbf{\textcolor{red}{0.379}} & \textbf{\textcolor{red}{0.410}} & 0.601 & 0.488 & \textbf{\textcolor{red}{0.463}} & \textbf{\textcolor{red}{0.443}} \\
    \bottomrule
    \end{tabular}%
    }
  \label{tab:main}%
\end{table*}

\paragraph{Zero-shot correction across TSFMs.}
Table~\ref{tab:main} reports, for each of the four TSFMs and each dataset, the
zero-shot forecast of the frozen model (\emph{Ori}) and the same forecast corrected
by \sysname\ (\emph{+\sysname}), averaged over the four horizons. \sysname\ improves
both metrics in all $56$ dataset--TSFM--metric cells of the table: every one of the
four TSFMs is improved on every one of the seven datasets, in MSE and in MAE. The
largest reductions come from the weakest frozen model at the longest horizon and
reach $34.5\%$ of MSE and $16.7\%$ of MAE---Moirai on Electricity at $H=720$, where
MSE falls from $1.476$ to $0.966$ and MAE from $0.942$ to $0.785$---with $30.0\%$ of
MSE under Sundial as well; fourteen of the $112$ settings remove at least a tenth of
the MSE, and ten at least a fifth. Complete per-setting results are in
Appendix~\ref{app:full-results}.

\paragraph{Correction after parameter-efficient adaptation.}
Table~\ref{tab:finetune} asks whether \sysname\ remains useful once the TSFM has
been adapted. Each TSFM is fine-tuned with LoRA on chronological training
windows and gets its own AMM and PGM; \emph{FT} denotes the fine-tuned model and
\emph{+\sysname} the same model corrected, with MSE and MAE averaged over the
four horizons. The correction still improves both metrics on all seven datasets
under Moirai and on six of them under Chronos and Sundial, again most where the
adapted model is weakest: fine-tuned Moirai on ETTh1 at $H=720$ goes from 1.077 to
0.753 MSE, a $30.1\%$ reduction. On TimesFM the correction is nearly switched off
--- the adapted checkpoints already track the target, so the gate suppresses the
correction and the average metrics move by less than $0.7\%$, the intended
behaviour of an evidence-driven correction rather than a failure. Complete results
are in Appendix~\ref{app:finetune}.

\begin{table}[t]
  \centering
  \caption{Forecasting with LoRA fine-tuned TSFMs.  For each TSFM,
  \emph{FT} is the LoRA fine-tuned model and \emph{+\sysname} the same
  fine-tuned model corrected by \sysname; MSE and MAE are averaged over
  $H\in\{96,192,336,720\}$, and \textbf{\textcolor{red}{red}} marks the better of the two
  within each TSFM--metric pair.  Lower is better.}
  \resizebox{\columnwidth}{!}{%
    \begin{tabular}{c|cccc|cccc|cccc|cccc}
    \toprule
    \multirow{2}[4]{*}{Models} & \multicolumn{4}{c|}{\textbf{Sundial}} & \multicolumn{4}{c|}{\textbf{Chronos}} & \multicolumn{4}{c|}{\textbf{TimesFM}} & \multicolumn{4}{c}{\textbf{Moirai}} \\
\cmidrule{2-17}          & \multicolumn{2}{c}{FT} & \multicolumn{2}{c|}{+\sysname\ (Ours)} & \multicolumn{2}{c}{FT} & \multicolumn{2}{c|}{+\sysname\ (Ours)} & \multicolumn{2}{c}{FT} & \multicolumn{2}{c|}{+\sysname\ (Ours)} & \multicolumn{2}{c}{FT} & \multicolumn{2}{c}{+\sysname\ (Ours)} \\
    \midrule
    \multicolumn{1}{c|}{Metric} & MSE & MAE & MSE & MAE & MSE & MAE & MSE & MAE & MSE & MAE & MSE & MAE & MSE & MAE & MSE & MAE \\
    \midrule
    ETTh1   & \textbf{\textcolor{red}{0.403}} & \textbf{\textcolor{red}{0.415}} & 0.411 & 0.417 & 0.461 & \textbf{\textcolor{red}{0.421}} & \textbf{\textcolor{red}{0.452}} & 0.422 & 0.489 & 0.438 & \textbf{\textcolor{red}{0.489}} & \textbf{\textcolor{red}{0.438}} & 0.594 & 0.493 & \textbf{\textcolor{red}{0.505}} & \textbf{\textcolor{red}{0.462}} \\
    \midrule
    ETTh2   & 0.338 & 0.376 & \textbf{\textcolor{red}{0.335}} & \textbf{\textcolor{red}{0.374}} & 0.375 & 0.384 & \textbf{\textcolor{red}{0.370}} & \textbf{\textcolor{red}{0.383}} & 0.366 & 0.388 & \textbf{\textcolor{red}{0.365}} & \textbf{\textcolor{red}{0.387}} & 0.432 & 0.422 & \textbf{\textcolor{red}{0.394}} & \textbf{\textcolor{red}{0.404}} \\
    \midrule
    ETTm1   & 0.351 & 0.362 & \textbf{\textcolor{red}{0.343}} & \textbf{\textcolor{red}{0.361}} & 0.349 & 0.354 & \textbf{\textcolor{red}{0.341}} & \textbf{\textcolor{red}{0.352}} & 0.304 & 0.355 & \textbf{\textcolor{red}{0.303}} & \textbf{\textcolor{red}{0.354}} & 0.566 & 0.451 & \textbf{\textcolor{red}{0.477}} & \textbf{\textcolor{red}{0.422}} \\
    \midrule
    ETTm2   & 0.274 & 0.319 & \textbf{\textcolor{red}{0.262}} & \textbf{\textcolor{red}{0.311}} & 0.275 & 0.308 & \textbf{\textcolor{red}{0.274}} & \textbf{\textcolor{red}{0.307}} & 0.269 & 0.305 & \textbf{\textcolor{red}{0.268}} & \textbf{\textcolor{red}{0.305}} & 0.318 & 0.342 & \textbf{\textcolor{red}{0.293}} & \textbf{\textcolor{red}{0.329}} \\
    \midrule
    Weather & 0.249 & 0.279 & \textbf{\textcolor{red}{0.235}} & \textbf{\textcolor{red}{0.269}} & 0.236 & 0.247 & \textbf{\textcolor{red}{0.229}} & \textbf{\textcolor{red}{0.245}} & 0.229 & 0.250 & \textbf{\textcolor{red}{0.228}} & \textbf{\textcolor{red}{0.250}} & 0.280 & 0.288 & \textbf{\textcolor{red}{0.256}} & \textbf{\textcolor{red}{0.275}} \\
    \midrule
    Exchange& 0.496 & 0.462 & \textbf{\textcolor{red}{0.432}} & \textbf{\textcolor{red}{0.432}} & 0.396 & 0.420 & \textbf{\textcolor{red}{0.375}} & \textbf{\textcolor{red}{0.407}} & \textbf{\textcolor{red}{0.400}} & \textbf{\textcolor{red}{0.423}} & 0.426 & 0.435 & 0.599 & 0.487 & \textbf{\textcolor{red}{0.470}} & \textbf{\textcolor{red}{0.445}} \\
    \midrule
    ECL     & 0.146 & 0.242 & \textbf{\textcolor{red}{0.146}} & \textbf{\textcolor{red}{0.242}} & 0.157 & 0.243 & \textbf{\textcolor{red}{0.157}} & \textbf{\textcolor{red}{0.243}} & 0.159 & 0.246 & \textbf{\textcolor{red}{0.159}} & \textbf{\textcolor{red}{0.246}} & 0.467 & 0.430 & \textbf{\textcolor{red}{0.376}} & \textbf{\textcolor{red}{0.402}} \\
    \bottomrule
    \end{tabular}%
  }
  \label{tab:finetune}%
\end{table}

\subsection{Analysis}
\label{sec:exp-analysis}

\paragraph{Ablation of the core components.}
Table~\ref{tab:ablation} reports absolute MSE and MAE at $H=96$, averaged over
the four TSFMs, for the complete method and for five variants that either remove
one component or replace one stored quantity. Every component is needed: the
complete method is the best of the six rows in all $14$ dataset--metric columns,
and every variant is worse on the average of both metrics, the two replacements
losing the most. Appendix~\ref{app:ablations} breaks the variants down dataset by
dataset, and Appendix~\ref{app:comparison} compares \sysname\ with a
retrieval-augmented method that reuses the futures of similar windows.

\begin{table}[t]
  \centering
  \caption{Ablation at $H=96$: absolute MSE / MAE, averaged over the four TSFMs
  and the seven datasets. The first three rows remove one module, the last two
  replace the stored quantity or the recent level; \textbf{\textcolor{red}{red}}
  marks the best average.}
  \label{tab:ablation}
  \resizebox{\columnwidth}{!}{%
    \begin{tabular}{lcccccccccccccc}
    \toprule
    \multirow{2}{*}{Variants} & \multicolumn{2}{c}{ETTh1} & \multicolumn{2}{c}{ETTh2} & \multicolumn{2}{c}{ETTm1} & \multicolumn{2}{c}{ETTm2} & \multicolumn{2}{c}{Weather} & \multicolumn{2}{c}{Exchange} & \multicolumn{2}{c}{ECL} \\
\cmidrule{2-15}
     & mse & mae & mse & mae & mse & mae & mse & mae & mse & mae & mse & mae & mse & mae \\
    \midrule
w/o shape relation & 0.412 & 0.421 & 0.318 & 0.362 & 0.378 & 0.371 & 0.196 & 0.268 & 0.192 & 0.223 & 0.101 & 0.227 & 0.149 & 0.243 \\
w/o bias relation & 0.427 & 0.411 & 0.331 & 0.353 & 0.362 & 0.389 & 0.205 & 0.276 & 0.185 & 0.212 & 0.098 & 0.234 & 0.142 & 0.252 \\
w/o anchor & 0.403 & 0.432 & 0.324 & 0.341 & 0.387 & 0.365 & 0.191 & 0.258 & 0.201 & 0.225 & 0.104 & 0.218 & 0.153 & 0.237 \\
    \midrule
amplitude $\rightarrow$ mean residuals & 0.438 & 0.431 & 0.327 & 0.378 & 0.396 & 0.382 & 0.212 & 0.285 & 0.194 & 0.235 & 0.113 & 0.245 & 0.147 & 0.241 \\
recent level $\rightarrow$ naive ratio & 0.409 & 0.413 & 0.342 & 0.392 & 0.371 & 0.360 & 0.214 & 0.295 & 0.189 & 0.221 & 0.125 & 0.258 & 0.151 & 0.253 \\
    \midrule
\textbf{\sysname\ (full)} & \textbf{\textcolor{red}{0.381}} & \textbf{\textcolor{red}{0.387}} & \textbf{\textcolor{red}{0.295}} & \textbf{\textcolor{red}{0.335}} & \textbf{\textcolor{red}{0.344}} & \textbf{\textcolor{red}{0.349}} & \textbf{\textcolor{red}{0.183}} & \textbf{\textcolor{red}{0.254}} & \textbf{\textcolor{red}{0.179}} & \textbf{\textcolor{red}{0.209}} & \textbf{\textcolor{red}{0.093}} & \textbf{\textcolor{red}{0.212}} & \textbf{\textcolor{red}{0.138}} & \textbf{\textcolor{red}{0.227}} \\
    \bottomrule
    \end{tabular}%
  }
\end{table}

\paragraph{Reliability of the correction.}
The gate and the two fixed weights of Eq.~\eqref{eq:ratio} trade peak gain for
coverage: the same correction without the graph, applying the pooled retrieval
answer alone, improves both metrics in only $74$ of the $112$ settings and can lose
up to $8.4\%$ of MSE, whereas the deployed configuration improves both metrics in
$107$ of the $112$ settings. The correction is therefore the conservative of the
two, and the one whose gains hold in $33$ more settings.

\begin{figure}[t]
\centering
\includegraphics[width=0.99\textwidth]{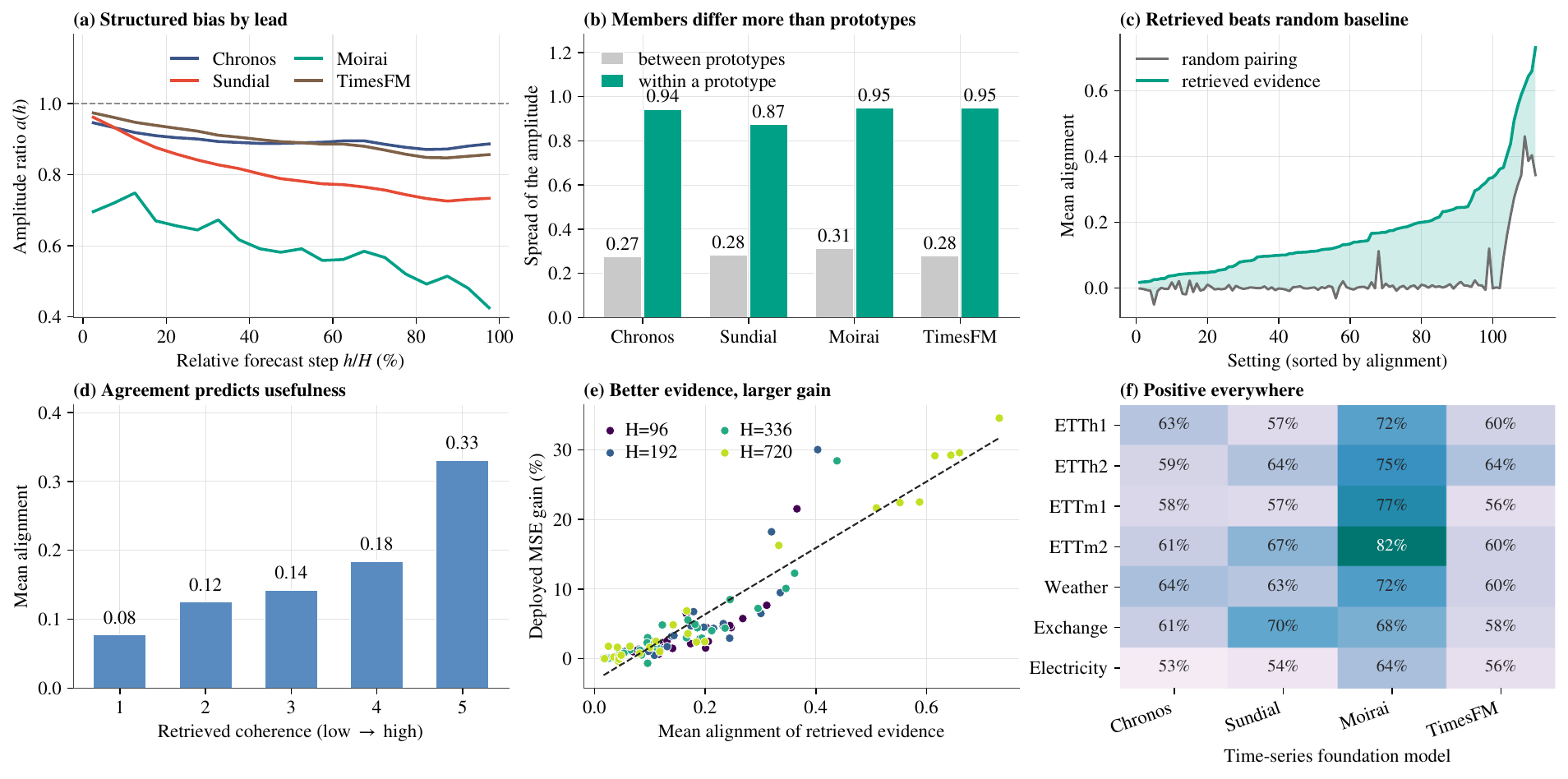}
\caption{Diagnostics of the pattern-conditioned bias over all 112 settings.
Panels (a)--(b) characterise the bias the memory stores, (c)--(e) the evidence
retrieval returns and its link to the gain, and (f) the coverage across
datasets and TSFMs.}
\label{fig:residual-diagnostics}
\end{figure}

\paragraph{Diagnostics of the pattern-conditioned bias.}
Figure~\ref{fig:residual-diagnostics} examines the bias over all 112 settings and
yields three insights. First, the bias is real and pattern-conditioned: the
frozen TSFMs systematically overstate how far the series departs from its recent
level, and that overstatement varies about three times more inside a prototype
than between prototypes, so the nearest pattern does not determine the bias of a
window and the memory has to aggregate related prototypes. Second, the retrieved
evidence is useful: the alignment between the correction it implies and the
realised error beats a random-pairing baseline in all $112$ settings and predicts
the per-setting MSE gain of Table~\ref{tab:main} with a correlation of $0.93$.
Third, the evidence is available almost everywhere, with $53\%$ to $82\%$ of
windows positively aligned depending on dataset and TSFM.


The remaining analyses, including the comparison with a retrieval-augmented
method that reuses futures and the efficiency measurements, are provided in
Appendix~\ref{app:comparison} and Appendix~\ref{app:efficiency}.

\section{Conclusion}
\label{sec:conclusion}

\sysname\ corrects the forecasts of a frozen time-series foundation model by retrieving, from the target history, the amplitude of the error that the model itself leaves on recurring patterns, which keeps the evidence dimensionless and comparable across windows. Across seven benchmarks, four horizons and four heterogeneous frozen models it improves both metrics on every dataset under all four TSFMs, reaching reductions of 34.5\% in MSE and 16.7\% in MAE on the weakest model at the longest horizon, and it adds only 7.7 to 12.2 microseconds per window on top of the frozen pipeline. Two directions follow naturally: updating the memory online as new windows arrive, and extending the bias retrieval to multivariate settings.

\bibliographystyle{iclr2027_conference}
\bibliography{references}

\clearpage
\appendix

\section{Experimental Details}
\label{app:details}
\label{app:reproducibility}

\subsection{Datasets and evaluation protocol}
\label{app:datasets}

\begin{table}[t]
\centering
\scriptsize
\caption{Benchmark characteristics and splitting protocol. The proportions are
train/validation/test, are the published ones and are chronological. The
memory-construction and calibration portions are cut from the training portion,
and the reported numbers are computed on the test portion.}
\label{tab:datasets}
\setlength{\tabcolsep}{3pt}
\begin{tabular}{lccc c}
\toprule
Dataset & Domain & Frequency & Channels & Split (train/val/test) \\
\midrule
ETTh1 & Electricity transformer & Hourly & 7 & 6:2:2 \\
ETTh2 & Electricity transformer & Hourly & 7 & 6:2:2 \\
ETTm1 & Electricity transformer & 15 minutes & 7 & 6:2:2 \\
ETTm2 & Electricity transformer & 15 minutes & 7 & 6:2:2 \\
Electricity & Household consumption & Hourly & 321 & 7:1:2 \\
Weather & Meteorology & 10 minutes & 21 & 7:1:2 \\
Exchange Rate & Foreign exchange & Daily & 8 & 7:1:2 \\
\bottomrule
\end{tabular}
\end{table}

Table~\ref{tab:datasets} lists the seven benchmarks and their split proportions.
All of them are used exactly as published, with their canonical
chronological splits~\cite{lai2018lstnet,wu2021autoformer,nie2023patchtst,zeng2023dlinear,liu2024itransformer,wang2024timemixer,qiu2024tfb,godahewa2021monash,aksu2024gifteval,alexandrov2020gluonts}, and both the memory-construction and the calibration portions are cut from
the training portion, separated from each other and from the published test
portion by a gap equal to the forecast horizon, so that no future observation of
one portion can enter a window of the next; the test portion is never touched
before evaluation. A separate memory is built for every dataset, horizon and TSFM,
while the retrieval encoder and the graph fusion module are shared across all of
them, and each channel is forecast independently.

\subsection{Metrics}
\label{app:metrics}

We report mean squared error (MSE) and mean absolute error (MAE), the two metrics
used by previous studies on these benchmarks. Both are computed per channel
against the un-normalised ground truth and then averaged over channels, and results
are reported per dataset and horizon; when a table reports a single number for a
dataset, as Table~\ref{tab:main} and Table~\ref{tab:comparison} do, it is the mean
of the four horizon values of that dataset.

\subsection{Memory construction and read-out settings}
\label{app:memory}

The offline phase has no optimisation: one pass over the memory-construction
portion of a target cell writes the record of
Section~\ref{sec:method-offline-prototype}, whose fields are the ones
Eq.~\eqref{eq:amp} to Eq.~\eqref{eq:coherence} define. Its principal settings are
$C=128$ prototypes, a recent level of $S=24$ observations, a member weight of
$|\mathrm{exc}|$, a support floor of $8$ for bias edges and $L=2$ neighbours per
relation. Online, the $K=3$ closest prototypes seed a subgraph of about $21$ nodes,
each expanded by three neighbours per relation. The read-out has width $96$, two
message-passing layers, four attention heads and dropout $0.2$, and enters the
correction through the anchor weight $\mu=0.35$ and the residual weight
$\nu=0.15$.

The read-out is fitted once, for $1500$ steps with AdamW and a cosine schedule and
three seeds, on the LOTSA corpus~\cite{woo2024moirai}, a multi-domain pretraining
corpus unrelated to the seven evaluation benchmarks, and is then frozen; no series,
forecast or label of an evaluation dataset takes part in that fitting, the frozen
TSFM and the retrieval encoder are never updated, and the memory of a target cell
is built only from the observed training portion of that cell. \sysname\ is
therefore a zero-shot method on the evaluation benchmarks: \emph{zero-shot}
throughout this paper means that the frozen TSFM is used as published, the
correction is fitted on an unrelated corpus, and no evaluation-domain data of any
kind enters training. All measurements are taken on a single H200 with batches of
$4096$ windows, and Table~\ref{tab:efficiency} reports the resulting cost of
$7.7$--$12.2$ microseconds per window.

\subsection{Retrieval and the graph fusion module}
\label{app:gfm}

At inference the query embedding is compared with the $C$ centroids and the
subgraph of Section~\ref{sec:method-online-retrieval} is assembled from the seeds,
the corpus prior and the anchor node: $K=3$ seeds, each expanded by three
neighbours per relation, which yields $21$ nodes on average. The record of a node
is the five-tuple of Section~\ref{sec:method-offline-prototype}, extended by the
log support relative to the median prototype, the coherence, the retrieval
similarity and the support flag; the query feature is the scale-normalised
excursion of Eq.~\eqref{eq:exc-res} followed by its log scale, its standard
deviation and its maximum. The graph fusion module keeps the three blocks of
Section~\ref{sec:method-online-fusion} at width $96$, $4$ attention heads and
dropout $0.2$~\cite{srivastava2014dropout}. Eq.~\eqref{eq:ceiling} is implemented
as written: each term is an attention-weighted average over the subgraph, the four
gains $\kappa_i=\mathrm{softplus}(\cdot)$ start at $\log 2$, the bias $b$ starts at
$2$, and a subgraph whose nodes disagree about the scaling cannot be capped
permissively. The deployed ratio is the one of Eq.~\eqref{eq:ratio}, with the
anchor weight $\mu=0.35$ and the residual weight $\nu=0.15$; the gain head's last
layer starts at zero so that $g\equiv1$ and training begins from the fixed terms
plus the vote, while the gate begins permissive and is only trained down where the
evidence is thin. The head is fitted once on the LOTSA
corpus~\cite{woo2024moirai}, whose series are split into memory and calibration
portions with a fixed seed, and is then frozen for all target cells; the frozen
TSFM and the retrieval encoder are never updated, and no label of an evaluation
domain is used at any point. The fitting objective is the error of the corrected
forecast---a squared and an absolute term---plus a penalty that keeps the returned
amplitude close to the pooled retrieval answer.

\subsection{Baselines and fine-tuning protocol}
\label{app:baselines}

The baseline throughout is the frozen TSFM itself, whose zero-shot forecast is
compared with the same forecast corrected by \sysname, on the same dataset and
horizon and with the same metric. The parameter-efficient baselines are LoRA
adapters optimised~\cite{kingma2015adam} with each TSFM's native
training loss on chronological
training windows; the adapter strength and the checkpoint are selected on a later
chronological tail of the training split, and the selected model is frozen before
the memory is built, so that the fine-tuned setting follows the same contract as
the zero-shot one.

\subsection{Protocol for the diagnostics}
\label{app:diag-protocol}

The diagnostics of Figure~\ref{fig:residual-diagnostics} cover all 112 settings.
To keep the pass cheap, each cell contributes at most 2000 test windows drawn
uniformly over the test portion, and the forecast steps are thinned to 96
uniformly spaced leads before horizons are pooled on the relative axis $h/H$;
alignments, spreads and counts are computed on that thinned grid, and the
amplitude of a window is its $|\mathrm{exc}|$-weighted average. The
random-pairing baseline permutes the retrieved evidence across test windows and
averages five permutations. The quality of the retrieved evidence is measured by
the cosine between the correction it implies and the realised error, which the
main text calls its alignment.

\subsection{Compute}
\label{app:compute}

Building the memory is a single pass over the target history: $O(n \cdot C)$
comparisons for the assignment, then one pass per prototype to aggregate its
members. A query costs one comparison against the centroids, the assembly of a
21-node subgraph, and the forward pass of the graph fusion module, whose
parameters are three orders of magnitude below the frozen TSFM's. The memory
itself stores $C$ records of $H$ amplitude values plus five scalars, that is
$128\times H$ floating point numbers per dataset, horizon and TSFM, which is
negligible next to the frozen model.

\section{Experimental Results}
\label{app:results}

\subsection{Complete results over the 112 settings}
\label{app:full-results}

\begin{table*}[t]
  \centering
  \scriptsize
  \caption{Complete zero-shot results behind Table~\ref{tab:main}, every dataset and horizon. For each TSFM, \emph{Ori} is the bare frozen model and \emph{+\sysname} the corrected forecast; MSE is compared with MSE and MAE with MAE, and \textbf{\textcolor{red}{red}} marks the better of the two.}
  \resizebox{\textwidth}{!}{%
    \begin{tabular}{c|c|cccc|cccc|cccc|cccc}
    \toprule
    \multicolumn{2}{c|}{\multirow{2}[4]{*}{Models}} & \multicolumn{4}{c|}{\textbf{Sundial}} & \multicolumn{4}{c|}{\textbf{Chronos}} & \multicolumn{4}{c|}{\textbf{TimesFM}} & \multicolumn{4}{c}{\textbf{Moirai}} \\
    \cmidrule{3-18}
    \multicolumn{2}{c|}{} & \multicolumn{2}{c}{Ori} & \multicolumn{2}{c|}{+\sysname\ (Ours)} & \multicolumn{2}{c}{Ori} & \multicolumn{2}{c|}{+\sysname\ (Ours)} & \multicolumn{2}{c}{Ori} & \multicolumn{2}{c|}{+\sysname\ (Ours)} & \multicolumn{2}{c}{Ori} & \multicolumn{2}{c}{+\sysname\ (Ours)} \\
    \midrule
    \multicolumn{2}{c|}{Metric} & MSE & MAE & MSE & MAE & MSE & MAE & MSE & MAE & MSE & MAE & MSE & MAE & MSE & MAE & MSE & MAE \\
    \midrule
    \multirow{4}{*}{ETTh1}
     & 96 & 0.356 & 0.383 & \textbf{\textcolor{red}{0.352}} & \textbf{\textcolor{red}{0.381}} & 0.385 & 0.380 & \textbf{\textcolor{red}{0.376}} & \textbf{\textcolor{red}{0.378}} & 0.384 & 0.382 & \textbf{\textcolor{red}{0.380}} & \textbf{\textcolor{red}{0.381}} & 0.426 & 0.411 & \textbf{\textcolor{red}{0.416}} & \textbf{\textcolor{red}{0.408}} \\
     & 192 & 0.390 & 0.406 & \textbf{\textcolor{red}{0.384}} & \textbf{\textcolor{red}{0.403}} & 0.442 & 0.412 & \textbf{\textcolor{red}{0.427}} & \textbf{\textcolor{red}{0.409}} & 0.428 & 0.408 & \textbf{\textcolor{red}{0.421}} & \textbf{\textcolor{red}{0.406}} & 0.467 & 0.433 & \textbf{\textcolor{red}{0.454}} & \textbf{\textcolor{red}{0.429}} \\
     & 336 & 0.403 & 0.416 & \textbf{\textcolor{red}{0.400}} & \textbf{\textcolor{red}{0.415}} & 0.481 & 0.433 & \textbf{\textcolor{red}{0.467}} & \textbf{\textcolor{red}{0.430}} & 0.446 & 0.419 & \textbf{\textcolor{red}{0.441}} & \textbf{\textcolor{red}{0.418}} & 0.478 & 0.446 & \textbf{\textcolor{red}{0.463}} & \textbf{\textcolor{red}{0.443}} \\
     & 720 & 0.438 & 0.444 & \textbf{\textcolor{red}{0.420}} & \textbf{\textcolor{red}{0.433}} & 0.483 & 0.449 & \textbf{\textcolor{red}{0.475}} & \textbf{\textcolor{red}{0.439}} & \textbf{\textcolor{red}{0.428}} & 0.427 & 0.428 & \textbf{\textcolor{red}{0.423}} & 1.112 & 0.725 & \textbf{\textcolor{red}{0.787}} & \textbf{\textcolor{red}{0.612}} \\
    \midrule
    \multirow{4}{*}{ETTh2}
     & 96 & 0.279 & 0.334 & \textbf{\textcolor{red}{0.274}} & \textbf{\textcolor{red}{0.329}} & 0.290 & 0.325 & \textbf{\textcolor{red}{0.286}} & \textbf{\textcolor{red}{0.324}} & 0.306 & 0.338 & \textbf{\textcolor{red}{0.301}} & \textbf{\textcolor{red}{0.335}} & 0.332 & 0.361 & \textbf{\textcolor{red}{0.317}} & \textbf{\textcolor{red}{0.352}} \\
     & 192 & 0.342 & 0.376 & \textbf{\textcolor{red}{0.340}} & \textbf{\textcolor{red}{0.373}} & 0.360 & 0.370 & \textbf{\textcolor{red}{0.358}} & \textbf{\textcolor{red}{0.369}} & 0.376 & 0.384 & \textbf{\textcolor{red}{0.372}} & \textbf{\textcolor{red}{0.381}} & 0.413 & 0.411 & \textbf{\textcolor{red}{0.395}} & \textbf{\textcolor{red}{0.401}} \\
     & 336 & \textbf{\textcolor{red}{0.361}} & 0.395 & 0.363 & \textbf{\textcolor{red}{0.395}} & 0.403 & \textbf{\textcolor{red}{0.403}} & \textbf{\textcolor{red}{0.403}} & 0.403 & 0.401 & 0.409 & \textbf{\textcolor{red}{0.399}} & \textbf{\textcolor{red}{0.408}} & 0.445 & 0.438 & \textbf{\textcolor{red}{0.426}} & \textbf{\textcolor{red}{0.429}} \\
     & 720 & 0.392 & 0.426 & \textbf{\textcolor{red}{0.379}} & \textbf{\textcolor{red}{0.422}} & 0.414 & 0.421 & \textbf{\textcolor{red}{0.401}} & \textbf{\textcolor{red}{0.412}} & 0.405 & 0.422 & \textbf{\textcolor{red}{0.403}} & \textbf{\textcolor{red}{0.416}} & 0.615 & 0.541 & \textbf{\textcolor{red}{0.482}} & \textbf{\textcolor{red}{0.478}} \\
    \midrule
    \multirow{4}{*}{ETTm1}
     & 96 & 0.292 & 0.325 & \textbf{\textcolor{red}{0.288}} & \textbf{\textcolor{red}{0.325}} & 0.332 & 0.333 & \textbf{\textcolor{red}{0.323}} & \textbf{\textcolor{red}{0.332}} & 0.313 & 0.329 & \textbf{\textcolor{red}{0.310}} & \textbf{\textcolor{red}{0.329}} & 0.479 & 0.417 & \textbf{\textcolor{red}{0.457}} & \textbf{\textcolor{red}{0.410}} \\
     & 192 & 0.342 & 0.357 & \textbf{\textcolor{red}{0.336}} & \textbf{\textcolor{red}{0.356}} & 0.387 & 0.366 & \textbf{\textcolor{red}{0.377}} & \textbf{\textcolor{red}{0.364}} & 0.361 & 0.359 & \textbf{\textcolor{red}{0.357}} & \textbf{\textcolor{red}{0.359}} & 0.517 & 0.440 & \textbf{\textcolor{red}{0.491}} & \textbf{\textcolor{red}{0.432}} \\
     & 336 & 0.378 & 0.380 & \textbf{\textcolor{red}{0.371}} & \textbf{\textcolor{red}{0.379}} & 0.436 & 0.393 & \textbf{\textcolor{red}{0.426}} & \textbf{\textcolor{red}{0.392}} & 0.394 & 0.382 & \textbf{\textcolor{red}{0.390}} & \textbf{\textcolor{red}{0.382}} & 0.651 & 0.472 & \textbf{\textcolor{red}{0.596}} & \textbf{\textcolor{red}{0.464}} \\
     & 720 & 0.451 & 0.418 & \textbf{\textcolor{red}{0.443}} & \textbf{\textcolor{red}{0.408}} & 0.528 & 0.439 & \textbf{\textcolor{red}{0.518}} & \textbf{\textcolor{red}{0.428}} & 0.456 & 0.418 & \textbf{\textcolor{red}{0.454}} & \textbf{\textcolor{red}{0.410}} & 1.314 & 0.763 & \textbf{\textcolor{red}{0.925}} & \textbf{\textcolor{red}{0.633}} \\
    \midrule
    \multirow{4}{*}{ETTm2}
     & 96 & 0.176 & 0.255 & \textbf{\textcolor{red}{0.168}} & \textbf{\textcolor{red}{0.249}} & 0.178 & 0.245 & \textbf{\textcolor{red}{0.175}} & \textbf{\textcolor{red}{0.243}} & 0.179 & 0.246 & \textbf{\textcolor{red}{0.176}} & \textbf{\textcolor{red}{0.244}} & 0.232 & 0.291 & \textbf{\textcolor{red}{0.214}} & \textbf{\textcolor{red}{0.281}} \\
     & 192 & 0.239 & 0.298 & \textbf{\textcolor{red}{0.228}} & \textbf{\textcolor{red}{0.290}} & 0.251 & 0.294 & \textbf{\textcolor{red}{0.248}} & \textbf{\textcolor{red}{0.291}} & 0.249 & 0.293 & \textbf{\textcolor{red}{0.245}} & \textbf{\textcolor{red}{0.291}} & 0.333 & 0.344 & \textbf{\textcolor{red}{0.302}} & \textbf{\textcolor{red}{0.330}} \\
     & 336 & 0.297 & 0.336 & \textbf{\textcolor{red}{0.284}} & \textbf{\textcolor{red}{0.327}} & 0.317 & 0.336 & \textbf{\textcolor{red}{0.313}} & \textbf{\textcolor{red}{0.334}} & 0.314 & 0.336 & \textbf{\textcolor{red}{0.308}} & \textbf{\textcolor{red}{0.332}} & 0.448 & 0.402 & \textbf{\textcolor{red}{0.393}} & \textbf{\textcolor{red}{0.381}} \\
     & 720 & 0.388 & 0.393 & \textbf{\textcolor{red}{0.374}} & \textbf{\textcolor{red}{0.384}} & 0.416 & 0.395 & \textbf{\textcolor{red}{0.413}} & \textbf{\textcolor{red}{0.393}} & 0.420 & 0.396 & \textbf{\textcolor{red}{0.409}} & \textbf{\textcolor{red}{0.391}} & 0.601 & 0.514 & \textbf{\textcolor{red}{0.466}} & \textbf{\textcolor{red}{0.449}} \\
    \midrule
    \multirow{4}{*}{ECL}
     & 96 & 0.118 & 0.213 & \textbf{\textcolor{red}{0.117}} & \textbf{\textcolor{red}{0.213}} & 0.124 & 0.211 & \textbf{\textcolor{red}{0.124}} & \textbf{\textcolor{red}{0.211}} & 0.133 & 0.220 & \textbf{\textcolor{red}{0.132}} & \textbf{\textcolor{red}{0.220}} & 0.180 & 0.266 & \textbf{\textcolor{red}{0.179}} & \textbf{\textcolor{red}{0.266}} \\
     & 192 & 0.135 & 0.231 & \textbf{\textcolor{red}{0.134}} & \textbf{\textcolor{red}{0.230}} & 0.143 & 0.229 & \textbf{\textcolor{red}{0.143}} & \textbf{\textcolor{red}{0.229}} & 0.150 & 0.236 & \textbf{\textcolor{red}{0.149}} & \textbf{\textcolor{red}{0.236}} & 0.202 & 0.281 & \textbf{\textcolor{red}{0.200}} & \textbf{\textcolor{red}{0.280}} \\
     & 336 & 0.150 & 0.247 & \textbf{\textcolor{red}{0.150}} & \textbf{\textcolor{red}{0.247}} & 0.161 & 0.248 & \textbf{\textcolor{red}{0.161}} & \textbf{\textcolor{red}{0.248}} & 0.167 & 0.253 & \textbf{\textcolor{red}{0.167}} & \textbf{\textcolor{red}{0.253}} & 0.223 & 0.296 & \textbf{\textcolor{red}{0.220}} & \textbf{\textcolor{red}{0.295}} \\
     & 720 & 0.188 & 0.280 & \textbf{\textcolor{red}{0.188}} & \textbf{\textcolor{red}{0.280}} & 0.207 & 0.288 & \textbf{\textcolor{red}{0.206}} & \textbf{\textcolor{red}{0.288}} & 0.206 & 0.287 & \textbf{\textcolor{red}{0.206}} & \textbf{\textcolor{red}{0.287}} & 1.476 & 0.942 & \textbf{\textcolor{red}{0.966}} & \textbf{\textcolor{red}{0.785}} \\
    \midrule
    \multirow{4}{*}{Weather}
     & 96 & 0.160 & 0.207 & \textbf{\textcolor{red}{0.156}} & \textbf{\textcolor{red}{0.203}} & 0.178 & 0.210 & \textbf{\textcolor{red}{0.175}} & \textbf{\textcolor{red}{0.209}} & 0.163 & 0.194 & \textbf{\textcolor{red}{0.161}} & \textbf{\textcolor{red}{0.193}} & 0.240 & 0.231 & \textbf{\textcolor{red}{0.226}} & \textbf{\textcolor{red}{0.229}} \\
     & 192 & 0.213 & 0.255 & \textbf{\textcolor{red}{0.203}} & \textbf{\textcolor{red}{0.247}} & 0.224 & 0.254 & \textbf{\textcolor{red}{0.218}} & \textbf{\textcolor{red}{0.249}} & 0.216 & 0.242 & \textbf{\textcolor{red}{0.208}} & \textbf{\textcolor{red}{0.239}} & 0.351 & 0.281 & \textbf{\textcolor{red}{0.287}} & \textbf{\textcolor{red}{0.272}} \\
     & 336 & 0.267 & 0.295 & \textbf{\textcolor{red}{0.256}} & \textbf{\textcolor{red}{0.287}} & 0.284 & 0.295 & \textbf{\textcolor{red}{0.276}} & \textbf{\textcolor{red}{0.291}} & 0.289 & 0.289 & \textbf{\textcolor{red}{0.275}} & \textbf{\textcolor{red}{0.284}} & 0.357 & 0.340 & \textbf{\textcolor{red}{0.321}} & \textbf{\textcolor{red}{0.324}} \\
     & 720 & 0.336 & 0.344 & \textbf{\textcolor{red}{0.328}} & \textbf{\textcolor{red}{0.337}} & 0.367 & 0.347 & \textbf{\textcolor{red}{0.358}} & \textbf{\textcolor{red}{0.343}} & 0.385 & 0.345 & \textbf{\textcolor{red}{0.366}} & \textbf{\textcolor{red}{0.340}} & 0.549 & 0.479 & \textbf{\textcolor{red}{0.426}} & \textbf{\textcolor{red}{0.416}} \\
    \midrule
    \multirow{4}{*}{Exchange}
     & 96 & 0.132 & 0.249 & \textbf{\textcolor{red}{0.104}} & \textbf{\textcolor{red}{0.223}} & 0.099 & 0.219 & \textbf{\textcolor{red}{0.093}} & \textbf{\textcolor{red}{0.211}} & 0.089 & 0.208 & \textbf{\textcolor{red}{0.087}} & \textbf{\textcolor{red}{0.205}} & 0.093 & 0.213 & \textbf{\textcolor{red}{0.089}} & \textbf{\textcolor{red}{0.208}} \\
     & 192 & 0.291 & 0.380 & \textbf{\textcolor{red}{0.204}} & \textbf{\textcolor{red}{0.320}} & 0.194 & 0.314 & \textbf{\textcolor{red}{0.181}} & \textbf{\textcolor{red}{0.302}} & 0.191 & 0.312 & \textbf{\textcolor{red}{0.182}} & \textbf{\textcolor{red}{0.305}} & 0.198 & 0.314 & \textbf{\textcolor{red}{0.185}} & \textbf{\textcolor{red}{0.304}} \\
     & 336 & 0.544 & 0.535 & \textbf{\textcolor{red}{0.389}} & \textbf{\textcolor{red}{0.454}} & 0.342 & 0.424 & \textbf{\textcolor{red}{0.325}} & \textbf{\textcolor{red}{0.412}} & 0.351 & 0.431 & \textbf{\textcolor{red}{0.331}} & \textbf{\textcolor{red}{0.418}} & 0.368 & 0.434 & \textbf{\textcolor{red}{0.342}} & \textbf{\textcolor{red}{0.420}} \\
     & 720 & 1.265 & 0.830 & \textbf{\textcolor{red}{1.059}} & \textbf{\textcolor{red}{0.765}} & 0.847 & 0.690 & \textbf{\textcolor{red}{0.832}} & \textbf{\textcolor{red}{0.684}} & 0.984 & 0.731 & \textbf{\textcolor{red}{0.916}} & \textbf{\textcolor{red}{0.712}} & 1.744 & 0.993 & \textbf{\textcolor{red}{1.236}} & \textbf{\textcolor{red}{0.841}} \\
    \bottomrule
    \end{tabular}%
    }
  \label{tab:gain-matrix}%
\end{table*}

Table~\ref{tab:gain-matrix} lists every dataset, horizon and TSFM, and reads as
the settings behind Table~\ref{tab:main}: both metrics improve in $107$ of the
$112$ settings, the largest reductions reach $34.5\%$ of MSE and $16.7\%$ of MAE
(Moirai on Electricity at $H=720$, from $1.476$ to $0.966$ MSE), fourteen settings
remove at least a tenth of the MSE and ten at least a fifth, and the gains grow with
the horizon and are largest where the frozen model is weakest, which the four column
blocks make visible: the Moirai block is the largest of the four and the TimesFM
block the smallest.

\begin{figure}[t]
\centering
\includegraphics[width=\textwidth]{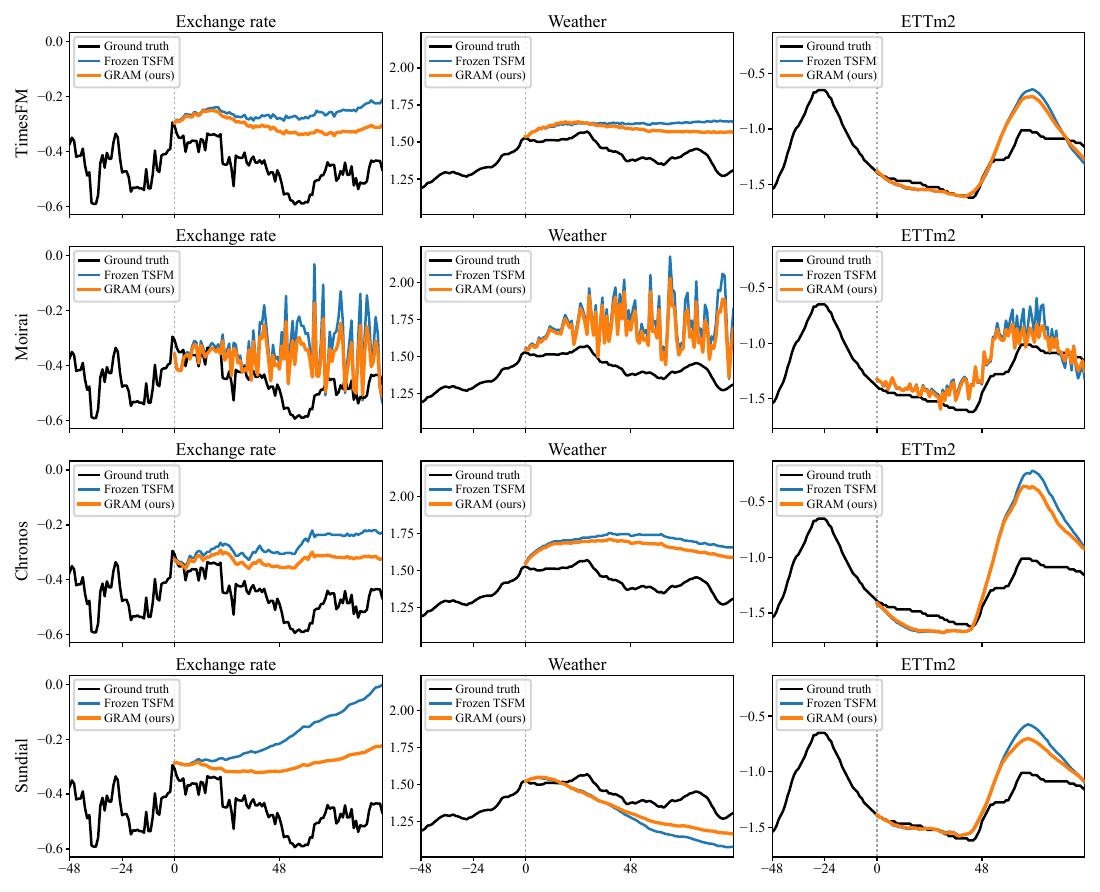}
\caption{Four frozen models (one per row, named on the left: TimesFM, Moirai, Chronos, Sundial) on three datasets (one per column) at $H=96$. Each panel is a single test window --- the same window for the rows of a column --- with the last $48$ steps of the series before the dotted line and the $96$-step forecast after it; the two forecasts are drawn only inside the forecast window, and all values are standardised. The windows are chosen so that every model is clearly off before the correction (frozen RMSE between $18\%$ and $55\%$ of the panel's own vertical range) and close after it (residual at most $42\%$ of that range, $+25\%$ to $+63\%$ of the squared error); among the windows that qualify, the one drawn is the window whose worst-case residual is the smallest. The Moirai row is the jitteriest of the four, which is the model's own behaviour on these series rather than the correction's.}
\label{fig:case-study}
\end{figure}

Figure~\ref{fig:case-study} is the window-level view at $H=96$: four frozen
models in the rows, three datasets in the columns, and one test window per
column. The windows are picked for the two properties a case study is meant to
show at once: before the correction every model is visibly off the label, and
after it the corrected forecast sits close to the label. Every panel removes
between $25\%$ and $63\%$ of the squared error, and in every panel the corrected
forecast is closer to the label than the frozen one at most forecast steps.

\subsection{Parameter-efficient adaptation}
\label{app:finetune}

\begin{table*}[t]
  \centering
  \scriptsize
  \caption{Complete results behind Table~\ref{tab:finetune}, every dataset and horizon. For each TSFM, \emph{FT} is the LoRA fine-tuned model and \emph{+\sysname} the same fine-tuned model corrected by \sysname; MSE is compared with MSE and MAE with MAE, and \textbf{\textcolor{red}{red}} marks the better of the two.}
  \resizebox{\textwidth}{!}{%
    \begin{tabular}{c|c|cccc|cccc|cccc|cccc}
    \toprule
    \multicolumn{2}{c|}{\multirow{2}[4]{*}{Models}} & \multicolumn{4}{c|}{\textbf{Sundial}} & \multicolumn{4}{c|}{\textbf{Chronos}} & \multicolumn{4}{c|}{\textbf{TimesFM}} & \multicolumn{4}{c}{\textbf{Moirai}} \\
    \cmidrule{3-18}
    \multicolumn{2}{c|}{} & \multicolumn{2}{c}{FT} & \multicolumn{2}{c|}{+\sysname\ (Ours)} & \multicolumn{2}{c}{FT} & \multicolumn{2}{c|}{+\sysname\ (Ours)} & \multicolumn{2}{c}{FT} & \multicolumn{2}{c|}{+\sysname\ (Ours)} & \multicolumn{2}{c}{FT} & \multicolumn{2}{c}{+\sysname\ (Ours)} \\
    \midrule
    \multicolumn{2}{c|}{Metric} & MSE & MAE & MSE & MAE & MSE & MAE & MSE & MAE & MSE & MAE & MSE & MAE & MSE & MAE & MSE & MAE \\
    \midrule
    \multirow{5}[2]{*}{ETTh1}
     & 96 & \textbf{\textcolor{red}{0.368}} & \textbf{\textcolor{red}{0.387}} & 0.371 & 0.388 & 0.411 & \textbf{\textcolor{red}{0.387}} & \textbf{\textcolor{red}{0.400}} & 0.387 & 0.414 & 0.411 & \textbf{\textcolor{red}{0.414}} & \textbf{\textcolor{red}{0.411}} & 0.410 & 0.403 & \textbf{\textcolor{red}{0.402}} & \textbf{\textcolor{red}{0.401}} \\
     & 192 & 0.394 & 0.408 & \textbf{\textcolor{red}{0.393}} & \textbf{\textcolor{red}{0.407}} & 0.461 & 0.416 & \textbf{\textcolor{red}{0.441}} & \textbf{\textcolor{red}{0.413}} & 0.518 & 0.451 & \textbf{\textcolor{red}{0.518}} & \textbf{\textcolor{red}{0.451}} & 0.437 & 0.420 & \textbf{\textcolor{red}{0.424}} & \textbf{\textcolor{red}{0.416}} \\
     & 336 & \textbf{\textcolor{red}{0.407}} & \textbf{\textcolor{red}{0.418}} & 0.410 & 0.418 & 0.488 & 0.434 & \textbf{\textcolor{red}{0.473}} & \textbf{\textcolor{red}{0.432}} & 0.602 & 0.479 & \textbf{\textcolor{red}{0.602}} & \textbf{\textcolor{red}{0.479}} & 0.454 & 0.433 & \textbf{\textcolor{red}{0.442}} & \textbf{\textcolor{red}{0.431}} \\
     & 720 & \textbf{\textcolor{red}{0.444}} & \textbf{\textcolor{red}{0.449}} & 0.470 & 0.455 & \textbf{\textcolor{red}{0.483}} & \textbf{\textcolor{red}{0.448}} & 0.492 & 0.455 & 0.421 & 0.409 & \textbf{\textcolor{red}{0.421}} & \textbf{\textcolor{red}{0.409}} & 1.077 & 0.715 & \textbf{\textcolor{red}{0.753}} & \textbf{\textcolor{red}{0.599}} \\
    \midrule
    \multirow{5}[2]{*}{ETTh2}
     & 96 & 0.267 & 0.323 & \textbf{\textcolor{red}{0.263}} & \textbf{\textcolor{red}{0.320}} & 0.281 & 0.320 & \textbf{\textcolor{red}{0.274}} & \textbf{\textcolor{red}{0.317}} & \textbf{\textcolor{red}{0.287}} & \textbf{\textcolor{red}{0.332}} & 0.288 & 0.332 & 0.311 & 0.343 & \textbf{\textcolor{red}{0.303}} & \textbf{\textcolor{red}{0.339}} \\
     & 192 & 0.336 & 0.369 & \textbf{\textcolor{red}{0.334}} & \textbf{\textcolor{red}{0.367}} & 0.361 & 0.372 & \textbf{\textcolor{red}{0.357}} & \textbf{\textcolor{red}{0.371}} & 0.371 & 0.385 & \textbf{\textcolor{red}{0.371}} & \textbf{\textcolor{red}{0.384}} & 0.395 & 0.394 & \textbf{\textcolor{red}{0.389}} & \textbf{\textcolor{red}{0.391}} \\
     & 336 & \textbf{\textcolor{red}{0.362}} & \textbf{\textcolor{red}{0.393}} & 0.364 & 0.393 & 0.415 & \textbf{\textcolor{red}{0.411}} & \textbf{\textcolor{red}{0.412}} & 0.411 & 0.397 & 0.409 & \textbf{\textcolor{red}{0.396}} & \textbf{\textcolor{red}{0.408}} & 0.409 & 0.413 & \textbf{\textcolor{red}{0.404}} & \textbf{\textcolor{red}{0.411}} \\
     & 720 & 0.386 & 0.420 & \textbf{\textcolor{red}{0.382}} & \textbf{\textcolor{red}{0.417}} & 0.445 & 0.434 & \textbf{\textcolor{red}{0.437}} & \textbf{\textcolor{red}{0.433}} & 0.408 & 0.426 & \textbf{\textcolor{red}{0.407}} & \textbf{\textcolor{red}{0.425}} & 0.612 & 0.538 & \textbf{\textcolor{red}{0.480}} & \textbf{\textcolor{red}{0.476}} \\
    \midrule
    \multirow{5}[2]{*}{ETTm1}
     & 96 & 0.267 & 0.311 & \textbf{\textcolor{red}{0.262}} & \textbf{\textcolor{red}{0.311}} & 0.259 & 0.302 & \textbf{\textcolor{red}{0.255}} & \textbf{\textcolor{red}{0.301}} & 0.222 & 0.303 & \textbf{\textcolor{red}{0.220}} & \textbf{\textcolor{red}{0.302}} & 0.303 & 0.333 & \textbf{\textcolor{red}{0.296}} & \textbf{\textcolor{red}{0.330}} \\
     & 192 & 0.326 & 0.349 & \textbf{\textcolor{red}{0.316}} & \textbf{\textcolor{red}{0.347}} & 0.319 & 0.338 & \textbf{\textcolor{red}{0.312}} & \textbf{\textcolor{red}{0.336}} & 0.241 & 0.326 & \textbf{\textcolor{red}{0.234}} & \textbf{\textcolor{red}{0.323}} & 0.365 & 0.367 & \textbf{\textcolor{red}{0.355}} & \textbf{\textcolor{red}{0.365}} \\
     & 336 & 0.368 & 0.374 & \textbf{\textcolor{red}{0.359}} & \textbf{\textcolor{red}{0.373}} & 0.367 & 0.366 & \textbf{\textcolor{red}{0.359}} & \textbf{\textcolor{red}{0.364}} & \textbf{\textcolor{red}{0.262}} & 0.355 & 0.263 & \textbf{\textcolor{red}{0.355}} & 0.404 & 0.391 & \textbf{\textcolor{red}{0.392}} & \textbf{\textcolor{red}{0.388}} \\
     & 720 & 0.444 & 0.415 & \textbf{\textcolor{red}{0.436}} & \textbf{\textcolor{red}{0.413}} & 0.449 & 0.409 & \textbf{\textcolor{red}{0.437}} & \textbf{\textcolor{red}{0.407}} & \textbf{\textcolor{red}{0.491}} & \textbf{\textcolor{red}{0.434}} & 0.497 & 0.436 & 1.193 & 0.714 & \textbf{\textcolor{red}{0.865}} & \textbf{\textcolor{red}{0.607}} \\
    \midrule
    \multirow{5}[2]{*}{ETTm2}
     & 96 & 0.172 & 0.252 & \textbf{\textcolor{red}{0.165}} & \textbf{\textcolor{red}{0.245}} & 0.167 & 0.235 & \textbf{\textcolor{red}{0.166}} & \textbf{\textcolor{red}{0.234}} & \textbf{\textcolor{red}{0.165}} & 0.236 & 0.165 & \textbf{\textcolor{red}{0.236}} & 0.194 & 0.257 & \textbf{\textcolor{red}{0.185}} & \textbf{\textcolor{red}{0.253}} \\
     & 192 & 0.238 & 0.297 & \textbf{\textcolor{red}{0.227}} & \textbf{\textcolor{red}{0.289}} & 0.237 & 0.283 & \textbf{\textcolor{red}{0.237}} & \textbf{\textcolor{red}{0.282}} & 0.232 & 0.282 & \textbf{\textcolor{red}{0.231}} & \textbf{\textcolor{red}{0.282}} & 0.251 & 0.298 & \textbf{\textcolor{red}{0.243}} & \textbf{\textcolor{red}{0.293}} \\
     & 336 & 0.297 & 0.335 & \textbf{\textcolor{red}{0.283}} & \textbf{\textcolor{red}{0.326}} & 0.300 & 0.326 & \textbf{\textcolor{red}{0.300}} & \textbf{\textcolor{red}{0.325}} & 0.291 & 0.322 & \textbf{\textcolor{red}{0.290}} & \textbf{\textcolor{red}{0.321}} & 0.311 & 0.336 & \textbf{\textcolor{red}{0.301}} & \textbf{\textcolor{red}{0.332}} \\
     & 720 & 0.388 & 0.394 & \textbf{\textcolor{red}{0.374}} & \textbf{\textcolor{red}{0.385}} & 0.397 & 0.389 & \textbf{\textcolor{red}{0.395}} & \textbf{\textcolor{red}{0.387}} & 0.386 & 0.381 & \textbf{\textcolor{red}{0.385}} & \textbf{\textcolor{red}{0.380}} & 0.518 & 0.475 & \textbf{\textcolor{red}{0.444}} & \textbf{\textcolor{red}{0.437}} \\
    \midrule
    \multirow{5}[2]{*}{Weather}
     & 96 & 0.154 & 0.202 & \textbf{\textcolor{red}{0.148}} & \textbf{\textcolor{red}{0.196}} & 0.142 & 0.175 & \textbf{\textcolor{red}{0.140}} & \textbf{\textcolor{red}{0.174}} & 0.142 & 0.179 & \textbf{\textcolor{red}{0.141}} & \textbf{\textcolor{red}{0.178}} & 0.158 & 0.188 & \textbf{\textcolor{red}{0.155}} & \textbf{\textcolor{red}{0.187}} \\
     & 192 & 0.211 & 0.254 & \textbf{\textcolor{red}{0.200}} & \textbf{\textcolor{red}{0.245}} & 0.191 & 0.221 & \textbf{\textcolor{red}{0.188}} & \textbf{\textcolor{red}{0.220}} & 0.190 & 0.225 & \textbf{\textcolor{red}{0.190}} & \textbf{\textcolor{red}{0.225}} & 0.211 & 0.236 & \textbf{\textcolor{red}{0.207}} & \textbf{\textcolor{red}{0.234}} \\
     & 336 & 0.275 & 0.301 & \textbf{\textcolor{red}{0.256}} & \textbf{\textcolor{red}{0.288}} & 0.261 & 0.267 & \textbf{\textcolor{red}{0.250}} & \textbf{\textcolor{red}{0.264}} & 0.249 & 0.270 & \textbf{\textcolor{red}{0.248}} & \textbf{\textcolor{red}{0.269}} & 0.281 & 0.289 & \textbf{\textcolor{red}{0.273}} & \textbf{\textcolor{red}{0.285}} \\
     & 720 & 0.356 & 0.359 & \textbf{\textcolor{red}{0.335}} & \textbf{\textcolor{red}{0.346}} & 0.351 & 0.325 & \textbf{\textcolor{red}{0.336}} & \textbf{\textcolor{red}{0.322}} & 0.334 & 0.327 & \textbf{\textcolor{red}{0.332}} & \textbf{\textcolor{red}{0.326}} & 0.469 & 0.439 & \textbf{\textcolor{red}{0.390}} & \textbf{\textcolor{red}{0.395}} \\
    \midrule
    \multirow{5}[2]{*}{Exchange}
     & 96 & 0.105 & 0.225 & \textbf{\textcolor{red}{0.096}} & \textbf{\textcolor{red}{0.215}} & 0.101 & 0.219 & \textbf{\textcolor{red}{0.093}} & \textbf{\textcolor{red}{0.208}} & \textbf{\textcolor{red}{0.103}} & \textbf{\textcolor{red}{0.224}} & 0.109 & 0.231 & 0.093 & 0.212 & \textbf{\textcolor{red}{0.090}} & \textbf{\textcolor{red}{0.208}} \\
     & 192 & 0.227 & 0.335 & \textbf{\textcolor{red}{0.198}} & \textbf{\textcolor{red}{0.311}} & 0.205 & 0.319 & \textbf{\textcolor{red}{0.193}} & \textbf{\textcolor{red}{0.306}} & \textbf{\textcolor{red}{0.196}} & \textbf{\textcolor{red}{0.315}} & 0.207 & 0.323 & 0.197 & 0.313 & \textbf{\textcolor{red}{0.187}} & \textbf{\textcolor{red}{0.305}} \\
     & 336 & 0.442 & 0.481 & \textbf{\textcolor{red}{0.374}} & \textbf{\textcolor{red}{0.440}} & 0.369 & 0.433 & \textbf{\textcolor{red}{0.351}} & \textbf{\textcolor{red}{0.420}} & \textbf{\textcolor{red}{0.353}} & \textbf{\textcolor{red}{0.429}} & 0.377 & 0.444 & 0.377 & 0.439 & \textbf{\textcolor{red}{0.354}} & \textbf{\textcolor{red}{0.424}} \\
     & 720 & 1.211 & 0.808 & \textbf{\textcolor{red}{1.059}} & \textbf{\textcolor{red}{0.761}} & 0.908 & 0.710 & \textbf{\textcolor{red}{0.864}} & \textbf{\textcolor{red}{0.695}} & \textbf{\textcolor{red}{0.950}} & \textbf{\textcolor{red}{0.723}} & 1.011 & 0.741 & 1.728 & 0.983 & \textbf{\textcolor{red}{1.251}} & \textbf{\textcolor{red}{0.845}} \\
    \midrule
    \multirow{5}[2]{*}{ECL}
     & 96 & 0.116 & 0.212 & \textbf{\textcolor{red}{0.116}} & \textbf{\textcolor{red}{0.211}} & 0.122 & \textbf{\textcolor{red}{0.209}} & \textbf{\textcolor{red}{0.122}} & 0.209 & 0.127 & 0.216 & \textbf{\textcolor{red}{0.127}} & \textbf{\textcolor{red}{0.216}} & 0.171 & \textbf{\textcolor{red}{0.263}} & \textbf{\textcolor{red}{0.171}} & 0.263 \\
     & 192 & 0.133 & 0.230 & \textbf{\textcolor{red}{0.133}} & \textbf{\textcolor{red}{0.229}} & 0.141 & 0.227 & \textbf{\textcolor{red}{0.141}} & \textbf{\textcolor{red}{0.227}} & 0.144 & 0.232 & \textbf{\textcolor{red}{0.144}} & \textbf{\textcolor{red}{0.232}} & 0.189 & \textbf{\textcolor{red}{0.277}} & \textbf{\textcolor{red}{0.188}} & 0.277 \\
     & 336 & 0.149 & 0.247 & \textbf{\textcolor{red}{0.148}} & \textbf{\textcolor{red}{0.246}} & 0.160 & 0.247 & \textbf{\textcolor{red}{0.160}} & \textbf{\textcolor{red}{0.247}} & 0.162 & 0.251 & \textbf{\textcolor{red}{0.162}} & \textbf{\textcolor{red}{0.250}} & 0.205 & 0.292 & \textbf{\textcolor{red}{0.204}} & \textbf{\textcolor{red}{0.292}} \\
     & 720 & 0.188 & 0.280 & \textbf{\textcolor{red}{0.187}} & \textbf{\textcolor{red}{0.280}} & 0.205 & 0.287 & \textbf{\textcolor{red}{0.205}} & \textbf{\textcolor{red}{0.287}} & 0.203 & 0.287 & \textbf{\textcolor{red}{0.202}} & \textbf{\textcolor{red}{0.286}} & 1.301 & 0.888 & \textbf{\textcolor{red}{0.941}} & \textbf{\textcolor{red}{0.774}} \\
    \bottomrule
    \end{tabular}%
    }
  \label{tab:finetune-detail}%
\end{table*}

Training LoRA adapters on the target history shortens the distance the correction
has to cover, and \sysname\ still helps afterwards: averaged over the four
horizons, both metrics improve on all seven datasets under Moirai and on six of
them under Chronos and Sundial, and the reduction follows the pattern of the
zero-shot setting, largest where the adapted model is weakest.
TimesFM is the exception, at $-0.68\%$ MSE and $-0.29\%$ MAE: its adapted
checkpoints already track the target so closely that the gate suppresses the
correction, which is the intended behaviour rather than a failure. The largest
single gain is fine-tuned Moirai on ETTh1 at $H=720$, where the MSE falls from
$1.077$ to $0.753$, a $30.1\%$ reduction. Table~\ref{tab:finetune-detail} gives the
same comparison for every dataset and horizon.

\subsection{Component ablations}
\label{app:ablations}

Table~\ref{tab:ablation} removes one component at a time at $H=96$, averaged over
the four TSFMs and reported as the relative improvement over the frozen model in
each dataset. The complete method is the best of the six rows in all $14$
dataset--metric columns. Removing either relation of the prototype graph lowers both
averages ($3.47\%$ and $3.32\%$ MSE against $3.58\%$ for the full method), which
is the performance side of the structural result in
Figure~\ref{fig:residual-diagnostics}(b): the members of a prototype differ more
than the prototypes differ from each other, so evidence has to be pooled across
related prototypes. Removing the anchor also lowers both averages, because the
head then loses the retrieval baseline it can fall back on. Replacing the
amplitude curves by mean residual curves removes most of the benefit, leaving
$0.29\%$ MSE and $-0.91\%$ MAE, since borrowed residuals track the errors of their
own windows rather than those of the query. Replacing the recent level of the
query by the naive ratio between future and forecast performs worse than the
frozen model on average ($-0.90\%$ MSE, $-2.91\%$ MAE), and the columns it wins
come at the cost of the other metric: it leads the Weather column on MSE while
losing $2.40\%$ MAE there.

The two structural variants are close on the average, which is expected: removing
one relation of degree two leaves the other one to supply most of the
neighbourhood, and the per-dataset columns show where each relation carries the
larger share, as on ETTh2 and ETTm2 for the shape relation.

\subsection{Comparison with the state-of-the-art retrieval-augmented method}
\label{app:comparison}

\begin{table*}[t]
  \centering
  \scriptsize
  \caption{Comparison with the retrieval-augmented method that reuses the ground-truth futures of similar historical windows, TS-RAG~\cite{ning2025tsrag}, the state of the art of that family. Every entry is the average over the four horizons of the dataset; the better of the two methods is \textbf{\textcolor{red}{red}} in each column.}
  \label{tab:comparison}
  \setlength{\tabcolsep}{3pt}
  \begin{tabular}{ll|ccccccc}
    \toprule
    \multicolumn{2}{c|}{} & ETTh1 & ETTh2 & ETTm1 & ETTm2 & Weather & Exchange & ECL \\
    \midrule
    \multirow{2}{*}{TS-RAG} & MSE & 0.434 & 0.348 & 0.395 & 0.282 & 0.258 & \textbf{\textcolor{red}{0.332}} & 0.157 \\
     & MAE & 0.413 & \textbf{\textcolor{red}{0.375}} & 0.374 & 0.314 & 0.274 & \textbf{\textcolor{red}{0.392}} & 0.244 \\
    \midrule
    \multirow{2}{*}{\sysname\ (Ours)} & MSE & \textbf{\textcolor{red}{0.394}} & \textbf{\textcolor{red}{0.341}} & \textbf{\textcolor{red}{0.360}} & \textbf{\textcolor{red}{0.264}} & \textbf{\textcolor{red}{0.236}} & 0.439 & \textbf{\textcolor{red}{0.147}} \\
     & MAE & \textbf{\textcolor{red}{0.411}} & 0.380 & \textbf{\textcolor{red}{0.369}} & \textbf{\textcolor{red}{0.313}} & \textbf{\textcolor{red}{0.269}} & 0.440 & \textbf{\textcolor{red}{0.242}} \\
    \bottomrule
  \end{tabular}
\end{table*}

The related family closest to ours reuses the ground-truth futures of the
historical windows that are most similar to a query.
TS-RAG~\cite{ning2025tsrag}, the state of the art of that family, rewrites the
query with the futures of its nearest neighbours, and
Table~\ref{tab:comparison} reports it together with \sysname\ on the seven datasets
of Table~\ref{tab:main}; every entry is produced by the implementation of the
method it belongs to with the frozen model that method is built for, and each is
averaged over the four horizons of the dataset.

The two averages are close---MSE is $0.3115$ for \sysname\ against $0.3151$ for
TS-RAG, and MAE $0.3463$ against $0.3409$, both within two percent---so the
difference lies in what the two carry across, and there \sysname\ is the better of
the two on six of the seven datasets in MSE and on five in MAE, and over the $28$
individual settings it is the better one in $24$ of them in MSE and in $15$ in MAE.
TS-RAG is ahead on Exchange, the series closest to a random walk, which is the one
setting where reusing futures is the stronger choice. A reused future brings the recurring bias
together with the sample-specific part that the decomposition of
Section~\ref{sec:method} argues is not transferable, so its value depends on how
much of that part is shared between the retrieved window and the query; the
dimensionless ratio that \sysname\ transfers does not, which is why its placement
is the more even one across settings.

\subsection{Efficiency and memory budget}
\label{app:efficiency}

\begin{table}[t]
  \centering
  \small
  \caption{Efficiency of the correction, measured on an H200 with batches of
  $4096$ windows.  The read-out is the only trained part; the memory is what has
  to be stored next to the frozen backbone.  Retrieval is a comparison against
  $C=128$ centroids and stays below $0.01\,\mu$s per window, so the read-out
  dominates the measured cost.}
  \label{tab:efficiency}
  \setlength{\tabcolsep}{4pt}
  \begin{tabular}{lcccc}
  \toprule
   & $H=96$ & $H=192$ & $H=336$ & $H=720$ \\
  \midrule
  read-out parameters (M) & 0.26 & 0.32 & 0.40 & 0.63 \\
  prototype records (KB) & 98 & 194 & 338 & 722 \\
  \midrule
  read-out cost ($\mu$s / window) & 7.7 & 8.7 & 9.4 & 12.2 \\
  \bottomrule
  \end{tabular}
\end{table}

The read-out is the only trained part and stays below a million parameters at
every horizon, so the correction never dominates the cost of a backbone forward
pass: the frozen backbones carry two to three orders of magnitude more
parameters ($128$M--$2$B against $0.3$--$0.6$M). What has to be stored per
dataset, TSFM and horizon is the prototype memory---$C=128$ records of $H$
amplitudes and $H$ spreads plus a handful of scalars, the centroid of each record
being also the key it is retrieved by. Retrieval itself
is a comparison against $C$ centroids followed by the assembly of a 21-node
subgraph, and the head is a small graph network run once per window, so the
whole correction costs between $7.7$ and $12.2$ microseconds per window on an
H200.

\subsection{Limitations and future work}
\label{app:limitations}

The offline phase is a single pass over the target history, and the read-out is
fitted on the memory-construction and calibration portions before the test
stream starts, so both costs are one-off; amortising one read-out across the
four horizons instead of fitting each of them separately is a natural next step.
The prototype library is built from the target series itself, and the retrieval
pipeline itself is corpus-agnostic, so a memory that a frozen backbone can
arrive with---built on one corpus and reused on another---is the direction we
consider most promising. Our study covers seven
long-standing benchmarks, four foundation models and the four standard horizons,
all as deterministic point forecasts; probabilistic heads, other model families,
and higher-frequency or geographically distributed series are left to future
work.

\end{document}